\documentclass[lettersize,10pt,journal]{IEEEtran}
\usepackage{amsmath,amsfonts}
\usepackage{algorithm,algpseudocode}
\usepackage{threeparttable}
\usepackage{multicol}
\usepackage{multirow}
\usepackage{algorithmicx}
\usepackage{array}
\usepackage[caption=false,font=normalsize,labelfont=sf,textfont=sf]{subfig}
\usepackage{textcomp}
\usepackage{stfloats}
\usepackage{url}
\usepackage{verbatim}
\usepackage{graphicx}
\usepackage{cite}
\usepackage{xcolor}
\usepackage{bm}
\usepackage[colorinlistoftodos,prependcaption,textsize=tiny,textwidth=0.2\columnwidth]{todonotes}
\usepackage{booktabs} 
\usepackage{dsfont}
\usepackage{amsthm}
\usepackage{ifthen}
\newcommand{\rev}[1]{{\textcolor{black}{#1}}}
\usepackage{color,soul}
\usepackage{tabularx}
\newtheorem{claim}{Claim}
\usepackage{amsmath,algorithm,tabularx}
 
\makeatletter
\newcommand{\multiline}[1]{%
  \begin{tabularx}{\dimexpr\linewidth-\ALG@thistlm}[t]{@{}X@{}}
    #1
  \end{tabularx}
}
\makeatother

\newcommand{\myhl}{\textcolor{black}}

\definecolor{darkorchid}{rgb}{0.6, 0.2, 0.8}
\definecolor{darkpastelgreen}{rgb}{0.01, 0.75, 0.24}

\newcommand{\nobs}{N}
\newcommand{\realizationnobs}{n}

\usepackage{tabularx}

\begin{document}

\newboolean{hl}
\setboolean{hl}{false}

\title{On Collaboration in Distributed Parameter Estimation with Resource Constraints}

\author{
    Yu-Zhen Janice Chen, Daniel S. Menasch\'e, Don Towsley~\IEEEmembership{Fellow,~IEEE}
    \thanks{Manuscript received June 2023; revised May 2024.}
    \thanks{This research is partially supported by the Army Research Laboratory under Cooperative Agreement W911NF-17-2-0196 (IoBT CRA) and partially supported by CAPES, CNPq (315106/2023-9), and FAPERJ (E-26/201.376/2021 and E-26/010.002174/2019).}
    \thanks{Yu-Zhen Janice Chen and Don Towsley are with the  Manning College of Information and Computer Sciences at UMass Amherst. Daniel S. Menasché is with the Institute of Computing at UFRJ, Brazil. Email: \{yuzhenchen, towsley\}@cs.umass.edu, sadoc@ic.ufrj.br}
    \thanks{This paper has been accepted for publication in IEEE Transactions on Network and Service Management.
    	Copyright may be transferred without notice, after which this version may no longer be accessible.}
}



\maketitle

\begin{abstract}

\myhl{Effective resource allocation in sensor networks, IoT systems, and distributed computing is essential for applications such as environmental monitoring, surveillance, and smart infrastructure. Sensors or agents must optimize their resource allocation to maximize the accuracy of parameter estimation.} 
In this work, we consider a group of sensors \rev{or} agents, each sampling from \rev{a different variable} of a multivariate Gaussian distribution and having \rev{a different estimation objective}. We formulate a sensor \rev{or} agent's data collection and collaboration policy design problem as a Fisher information maximization (or Cramer-Rao bound minimization) problem.  \myhl{ This formulation captures a novel trade-off in energy use, between locally collecting univariate samples \rev{and} collaborating to produce multivariate samples.    
When knowledge of the correlation between variables is available, we analytically identify two \rev{cases}: (1) where the optimal data collection policy entails investing resources to transfer information for collaborative sampling, 
and (2) where knowledge of the correlation between samples cannot enhance estimation efficiency.}  
When knowledge of certain correlations is unavailable, but collaboration remains potentially beneficial, we propose novel approaches that apply multi-armed bandit algorithms to learn the optimal data collection and collaboration policy in our sequential distributed parameter estimation problem. We \rev{illustrate} the effectiveness of the proposed algorithms, \texttt{DOUBLE-F}, \texttt{DOUBLE-Z}, \texttt{UCB-F}, \texttt{UCB-Z}, through simulation.
\end{abstract}



\section{Introduction}

\IEEEPARstart{L}{earning} parameters of distributions is one of the fundamental problems in computer sciences and statistics. With the advance of IoT and Wireless Sensor Network \myhl{(WSN)} technologies, there has been emerging research interest in distributed~\cite{gao2022review, jordan2018communication, sharifnassab2019order, acharya2021interactive, barnes2019fisher, acharya2020inference} and/or sequential~\cite{willett2000good, wang2013joint, rad2010distributed, kar2011convergence, kar2012distributed, zhang2012distributed, jia2021resource, kar2014distributed, yan2013optimal} parameter estimation problems. 
\myhl{When it comes to sequentially and distributedly collecting data for estimation purposes in network settings, it is natural and practical to consider resource constraints~\cite{ribeiro2006bandwidth, willett2000good, baek2004minimizing,blatt2004distributed, koutsopoulos2014distributed}, e.g., energy or communication bandwidth constraints, as these resources could be costly and limited for many computation and communication systems.  }

\myhl{Sequential distributed parameter estimation~\cite{rad2010distributed, kar2011convergence, kar2012distributed, zhang2012distributed} often deals with vertically partitioned data, i.e., each sensor or agent \rev{observes} distinct features/dimensions of each data sample.
In network settings, sensors \rev{or} agents may have different modalities or may be 
\rev{geographically dispersed within the network}.
If the goal \rev{of} each sensor \rev{or} agent in the group is to estimate the full global parameter vector, then collaboration is \emph{necessary}  because the observations each agent obtains are only partially informative (vertically partitioned). Otherwise, if the goal of each sensor  \rev{or} agent is to estimate a local parameter governing its observations, then collaboration is potentially beneficial but \emph{optional}. In fact, in many IoT or WSN applications, it is not necessary for every sensor or agent to estimate the common parameter vector. For example, in applications such as change-point detection for manufacturing processes or smart agriculture management, each sensor may be responsible for monitoring a specific area and may only need to estimate parameters relevant to its assigned tasks.}

\IEEEpubidadjcol

\myhl{In this work, we study a sequential distributed parameter estimation problem with resource constraints and vertically partitioned data from the standpoint of an agent/sensor whose objective is to estimate a local unknown parameter \rev{from a multivariate distribution}.} 
The samples can be communicated 
among sensors or between geographically distributed sites, each with its own data processing/learning unit (hence, we refer to sensors and agents interchangeably in the rest of this paper). 
When constraints limit the amount of information that can be passed around, trade-offs arise between the quantity of data and the type of estimation that should be performed by the agents.

\myhl{To the best of our knowledge, we are the first to formulate the sequential distributed parameter estimation problem with vertically partitioned data \rev{under resource constraints} where an agent's objective is to estimate a local unknown parameter.
Our work differs from prior works~\cite{rad2010distributed, kar2011convergence, kar2012distributed, zhang2012distributed} where collaboration among agents is always necessary for each agent to estimate the entire parameter vector. \rev{In} our problem, each agent has a distinct learning objective and is able to independently learn about its parameter of interest.} Consequently, whether collaboration between agents is worthwhile becomes an intriguing problem.
Specifically, we pose the following research question: 
\begin{quote}
\emph{How should resource-constrained 
agents 
allocate their resources between individual learning (local data collection) 
and collaboration (data transfer 
from other
agents) in order to achieve the most accurate parameter estimation?}
\end{quote}

To answer this research question, we frame the problem of designing optimal data collection and collaboration policies/strategies as a problem of maximizing Fisher information (FI)~\cite{cover1999elements} and/or minimizing the Cramer-Rao bound (CRB), i.e., minimizing the reciprocal of the \rev{FI}~\cite{cover1999elements}. 
The \rev{FI}   quantifies the amount of information contained in a sample (such as a marginal sample or a bivariate/trivariate joint sample) concerning an unknown parameter. The \rev{CRB}  establishes a lower bound on the variance of unbiased estimators for that parameter, \rev{as a function of the FI}.

We study the optimization problem in three scenarios: \myhl{
\texttt{Scenario 1} where 
only the target parameter is unknown, \texttt{Scenario 2} where none of the mean parameters are known but the correlation information is available, and \texttt{Scenario 3} where the correlation information is unavailable and only the target parameter is unknown.} 
We investigate how the amount of correlation between data samples collected by distinct sensors affects the design of optimal data collection and collaboration policies.
Note that an agent always has the option to spend all its resource budget on local data collection and estimate the mean of its corresponding variable by computing the sample mean. We investigate the circumstances under which allocating a portion of the resource budget to collaboration (i.e., collecting data samples of additional variable(s) from other sensor(s)/agent(s) to form joint samples) provides more \rev{FI} and which estimators can be utilized correspondingly.

\begin{table*}[t]
	\centering
	\caption{Summary of results}
	\label{tab:summary} 
	\scalebox{0.95}{\myhl{
    \begin{tabular}{l|l|l|l}
    \toprule
     & \texttt{Scenario 1} & \texttt{Scenario 2} & \texttt{Scenario 3}\\
     \midrule
     Assumptions & Correlation Available & Correlation Available & Correlation Unavailable\\
     &Only Target Parameter Unknown & All Parameters Unknown & Only Target Parameter Unknown\\
     \midrule
     Knowledge of Correlation   & Useful & Not Useful & Useful, Need to Estimate\\
     \midrule
     To Collaborate or Not   & Yes & No & Yes\\
     \midrule
     Sampling \& Collaboration Policy & Static (Offline Optimization) & Static (Offline Optimization) & Adaptive (Multi-Armed Bandit-Based Algorithm)\\
     \midrule
     Estimator & Kalman Filter & Sample Mean & Kalman Filter\\
     \bottomrule
    \end{tabular}
    } }
\end{table*}

\subsection*{Our Contributions.}

\myhl{Our main findings and results are summarized in Table~\ref{tab:summary}.} Our contributions are as follows:

First, we develop a mathematical model for our sequential distributed parameter estimation problem that allows us to answer the question of what data is valuable to observe and transmit under resource constraints. 
In particular, when correlation information is available, our Fisher information model formulation allows us to analytically obtain closed-form optimization solutions and estimators \rev{that lead} to constructive data collection and collaboration policies; when certain correlation information is unavailable, our formulation allows us to design multi-armed bandit algorithms that learn the optimal data collection and collaboration policies.

Second, we provide optimal solutions for \texttt{Scenario 1} and \texttt{Scenario 2}. In \texttt{Scenario 1}, our analysis demonstrates that collaboration provides estimates with smaller variances. Specifically, we develop optimal static data collection policies as well as optimal estimators given different correlations, resource costs, and budgets. In \texttt{Scenario 2}, our analytical findings indicate that sample mean provides the optimal estimate of the mean parameter of interest, which \rev{suggests} collaboration is not beneficial in this scenario; even if correlation information is available, it cannot be leveraged. 

Third, we propose novel learning policies for \texttt{Scenario~3}, where the Fisher information optimization problem for estimating an unknown mean parameter cannot be solved in closed form because the \myhl{unknown} correlation parameters appear in the objective function. 
To tackle this problem, we adopt an iterative correlation coefficient estimation and data collection policy optimization strategy. We devise two surrogate rewards specific to our Fisher information policy optimization problem, enabling the application of multi-armed bandit algorithms to formulate adaptive data collection and collaboration policies. Our simulations show that the performance of our adaptive data collection strategies gets close to that of the optimal (oracle) static data collection policies over time.

We note that a preliminary version of this work appeared in~\cite{chen2021collaborate}. This full version here extends the bivariate results in~\cite{chen2021collaborate} to multivariate cases and provides constructive solutions for \texttt{Scenario 3} 
by tackling the corresponding sequential decision-making and exploration-exploitation trade-off problem.

\subsection*{Paper outline. }
The rest of the paper is organized as follows.    \myhl{Section~\ref{sec:related}    provides an extensive literature review.}  In Section~\ref{sec:prelim}, we present a mathematical formulation of our problem and formally introduce the three scenarios under study. 
\myhl{
In Sections~\ref{sec:correlation-known-1}, \ref{sec:correlation-known-2}, and \ref{sec:correlation-unknown}, we analyze \texttt{Scenario~1}, \texttt{Scenario 2}, and \texttt{Scenario 3} respectively, formulating the corresponding optimization problems, providing sampling and collaboration policies and estimators. 
}
Finally, in Section~\ref{sec:conclusion}, we conclude and discuss future directions.
Due to the page limit, we defer to appendices a notation table, derivations of some estimators, pseudo codes of proposed algorithms, and a discussion of potential applications that may benefit from our results.


\section{Related Work}\label{sec:related}

\rev{In this section, we report on related work pertaining to distributed parameter estimation (Section~\ref{sec:distributed_parameter_estimation}), the goals and collaboration strategies of agents (Section~\ref{sec:agents_goals_collaboration}), inference of Gaussian distribution parameters (Section~\ref{sec:gaussian_distribution_inference}), and experimental design leveraging Fisher Information (Section~\ref{sec:experimental_design_fisher_information}).}

\subsection{Distributed  parameter estimation} \label{sec:distributed_parameter_estimation}
Our study contributes to the broad field of distributed parameter estimation. Within this domain, considerable attention has been devoted to the offline setting, where each agent acquires multiple independent and identically distributed samples at the outset of the learning process~\cite{gao2022review, jordan2018communication, sharifnassab2019order, acharya2021interactive, barnes2019fisher, acharya2020inference}. Research in offline distributed estimation primarily focuses on determining the required number of bits for the estimation task. 

In contrast to the offline distributed estimation problem, the online distributed estimation problem addressed in this study involves agents estimating parameters while sequentially obtaining samples over time.
With the advances in Wireless Sensor Network applications, online (sequential) distributed estimation has attracted emerging research interests in recent years.   Specifically, two prominent research threads in this domain are distributed state estimation (a.k.a. filtering, data assimilation, data fusion)~\cite{jia2021resource, kar2014distributed, yan2013optimal} and distributed parameter estimation (a.k.a. inverse problem, system identification)~\cite{wang2013joint}, each characterized by distinct problem formulations and analysis methods. 
Distributed state estimation typically incorporates a state transition matrix or function in its model and aims to infer information about the system state at each time step based on measurements up to the current time. 
More closely related to our work is the distributed parameter estimation problem, \rev{where agents aim to estimate} a static parameter or parameter vector.

\myhl{Previous sequential distributed parameter estimation works~\cite{rad2010distributed, kar2011convergence, kar2012distributed, zhang2012distributed} typically assumed that all parameters are unknown and all agents must estimate all parameters.  Our work assumes that each agent aims at estimating a single local parameter and that other parameters may be known (\texttt{Scenario 1}), not known (\texttt{Scenario 2}), or partially known (\texttt{Scenario  3}). We are unaware of previous works that considered those three scenarios.}

\subsection{Agents goals and necessary versus  optional collaboration} \label{sec:agents_goals_collaboration}

\myhl{
In related  works~\cite{rad2010distributed, kar2011convergence, kar2012distributed, zhang2012distributed},  
the goal of each sensor/agent is to estimate a common unknown parameter vector. The observations each agent obtains are only partially informative (vertically partitioned) with respect to the common parameter vector of interest. 
Therefore, in those studies, agents typically need to collaborate with designated neighbors, as the full parameter vector is not locally identifiable.} 
 
\myhl{In contrast, in our work, agents decide whether collaboration is worthwhile.  
Specifically, in our setting, each agent can always estimate its parameter of interest via independent local data collection; it collaborates to obtain data from other agents only if it leads to more efficient estimation. }

It is also worth noting that the idea of collaboration considered in this work has a different flavor from that studied in the literature of combining estimation problems~\cite{efron1973combining}. Our focus lies in leveraging correlated data to produce more accurate unbiased estimates, while~\cite{efron1973combining} investigates seemingly unrelated data and possibly biased estimates.

Of particular relevance to this work is the literature on the inference of parameters of the Gaussian distribution. In particular, some of the early results on the amount of information contained in a sample with missing data, derived by Wilks~\cite{wilks1932moments} and later on extended by Bishwal and Pena~\cite{bishwal2008note}, serve as foundations for our search for optimal data collecting strategies accounting for maximum likelihood estimators.  
Whereas~\cite{wilks1932moments, bishwal2008note} assume that one has no control over missing data, in this paper, for designing optimal data collecting strategies, the goal is to determine which data must be ``missed", e.g., due to resource constraints.  By leveraging this observation,  we build on top of~\cite{wilks1932moments, bishwal2008note}, posing the design of collaborative estimation strategy as a constrained optimization problem and deriving 
its solution.

\subsection{Experimental design leveraging Fisher Information} \label{sec:gaussian_distribution_inference}

Our resource allocation problem shares similarities with the optimal experimental design problems~\cite{boyd2004convex}, which aim to determine the best setting of experimental conditions or data collection strategies to obtain the most informative and efficient data for statistical inference. 
Optimal design problems, with a rich history in statistics~\cite{silvey2013optimal}, have also been studied for many computer science problems, including sensor selection/placement~\cite{ucinski2007d, nagata2021data, joshi2008sensor}, network tomography~\cite{he2015fisher}, and network design problems~\cite{tsai2018efficient}. 
The formulation of our problem can be viewed as a relaxed version of the optimal design problem~\cite[Section 7.5.1]{boyd2004convex}, where the variables are relaxed from integers to fractions. Additionally, we further consider both static and adaptive versions of the optimal design problems and account for resource constraints.

Prior works have considered resource constraints in a variety of aspects, e.g., observation quantization requirements~\cite{ribeiro2006bandwidth, willett2000good}, or per-link constraints in a network~\cite{baek2004minimizing}. 
In this work,  \rev{we focus on the}  perspective of battery-powered IoT devices or WSN sensors, where onboard battery power is often limited and typically the critical resource. Hence, we formulate the resource constraints in a similar vein as~\cite{blatt2004distributed, koutsopoulos2014distributed} -- limited resource budgets hinder us from collecting and/or transmitting all the data.

One of the key elements in our problem formulation is the 
Fisher information, which allows us to analytically study the optimal static data collection and collaboration policies and design straightforward surrogate rewards for multi-armed bandit-based adaptive policies. Fisher information has been extensively considered in statistics fields~\cite{grambsch1983sequential, lane2020adaptive, lindsay1997second, cao2012relative} and
widely applied 
in the realm of computer networks, e.g., assessing the fundamental limits of flow size estimation~\cite{ribeiro2006fisher}, network tomography~\cite{he2015fisher} and sampling in sensor networks~\cite{song2009optimal, yilmaz2013sequential}. 
Our Fisher information-based methodology differs from previous work in at least two aspects. First, we assume that samples are collected from
a multivariate Gaussian distribution, which allows us to derive novel and provably optimal sampling strategies, some of which are amenable to closed-form expressions for the estimators. Second, Fisher information is not initially known in \texttt{Scenario 3}, and we propose to use multi-armed bandit (MAB) to balance the trade-off between estimating Fisher information and optimizing the acquired Fisher information, and we design two Fisher information-motivated surrogate rewards for our MAB-based adaptive policies.

\subsection{Sequential estimation and multi-armed bandits} \label{sec:experimental_design_fisher_information}

Sequential sampling, estimation, and testing have a rich history going back to the seminal works~\cite{wald1992sequential, chernoff1959sequential}. In particular, sequential correlation significance testing~\cite{kubinger2007testing, schneider2015sequential, rasch2015determination} is related to our iterative correlation estimation, data collection and collaboration strategy optimization process in \texttt{Scenario~3}. We resort to the MAB model, a fundamental reinforcement learning model~\cite{bubeck2012regret, tran2010epsilon, cayci2020budget, mo2022intelligent}, since the absence of correlation information makes it impossible to derive the optimal static collaboration strategy at the very beginning of the learning horizon. While most of the MAB research assumes that rewards will be given by the environment, there have also been recent works that study the MAB problem for applications whose rewards needed to be approximated due to, e.g., delayed response from phone application users~\cite{han2021budget} or having multiple possibly competing objective to balance with~\cite{mehrotra2020bandit}. \rev{The most relevant work to us is~\cite{masoumian2021sequential}, which also applies MAB-based algorithms for data sample collection; however, it considers horizontally partitioned data, whereas our study investigates vertically partitioned data.}


\section{Problem Formulation}\label{sec:prelim}

In the following, we first formulate the resource-constrained sensor/agent observation model \rev{(Section~\ref{sec:observationmodel})}. We then formally introduce the three scenarios, which differ according to the availability assumptions on parameters \rev{(Section~\ref{sec:distributed})}.  
Finally, we cast the resource-constrained data collection and collaboration problem as a multi-armed bandit problem and give mathematical descriptions of static and adaptive data collection and collaboration policies \rev{(Section~\ref{sec:datac})}. \myhl{Please refer to Appendix~\ref{sec:notation} for a table of notation.}

\subsection{Observation Model and Resource Constraints} \label{sec:observationmodel}
Consider a set of sensors/agents, $S_k$, $k \in [K]\equiv \{1,...,K\}$, which may or may not reside at the same location. We consider a time-slotted system, $t = 1,..., T$. At each time slot $t$, sensors $S_1, S_2, ..., S_K$ can each make an independent observation on random variables $X_1, X_2, ..., X_K$ respectively.  
Note that $X_1, X_2, ..., X_K$  can represent different modalities or features or the same modality but collected at different geographic locations. 
We model observations as coming from a multivariate Gaussian distribution with a mean vector ${\bm \mu} = (\mu_1, \mu_2, ..., \mu_K) \in \mathbb{R}^K$ and a covariance matrix\footnote{When we refer to the covariance matrix of a certain subset of variables, \rev{we list} the variables in subscript, e.g., ${\bm \Sigma}_{(X_1, X_k)}$ denotes the covariance of variables $X_1,X_k$, $k\in \{1,...,K\}$.} 
${\bm \Sigma} \in \mathbb{R}^{K\times K}$, whose $(k,\ell)$-th entry $\left({\bm \Sigma}\right)^{k,\ell} = \rho_{k, \ell} \sigma_k \sigma_{\ell}$ if $k \neq \ell$ and $\left({\bm \Sigma}\right)^{k,\ell} = \sigma_k^2$ if $k = \ell$, where $\sigma_k^{2}$ denotes the variance of $X_k$ and $\rho_{k,\ell}$ denotes the Pearson correlation coefficients of $X_k$ and $X_\ell$. 
Without loss of generality, we study the case where values of correlations are in $[0, 1)$. 
Henceforth, we refer to a single observation from a single sensor as a \emph{marginal observation} or simply as an \emph{observation}, 
whereas a \emph{joint observation} comprises two or more observations from two or more sensors at the same time slot. A sample refers to either a marginal or a joint observation.

Associated with each sensor is an agent capable of making
marginal observations, processing data samples locally, and transmitting/receiving data samples to/from other agents for collaborative estimation. 
There are resource costs associated with taking an observation and with transmitting it.  
We assume the cost to make an observation is one unit and that the cost of communication per observation (either to transmit or receive it) is $\alpha > 0$ unit(s).
For example, sensor $S_1$ locally collects a univariate sample $x_1$ at a cost of one unit of resource; if, in the meantime, $S_1$ also receives an observation, say $x_2$, to pair with $x_1$, then bivariate sample $(x_1, x_2)$ costs $S_1$ $\alpha+1$ units of resource; if $S_1$ receives two other observations, $x_2, x_3$, to pair with $x_1$, then trivariate sample $(x_1, x_2, x_3)$ costs $2\alpha+1$ units of resource. In addition, each sensor/agent is allocated a per time slot resource budget of $E$ (which is the same for all sensors), which may introduce a trade-off between collecting one multivariate sample and collecting more univariate samples.

\begin{figure}[t]
   \centering
    \includegraphics[width=\linewidth]{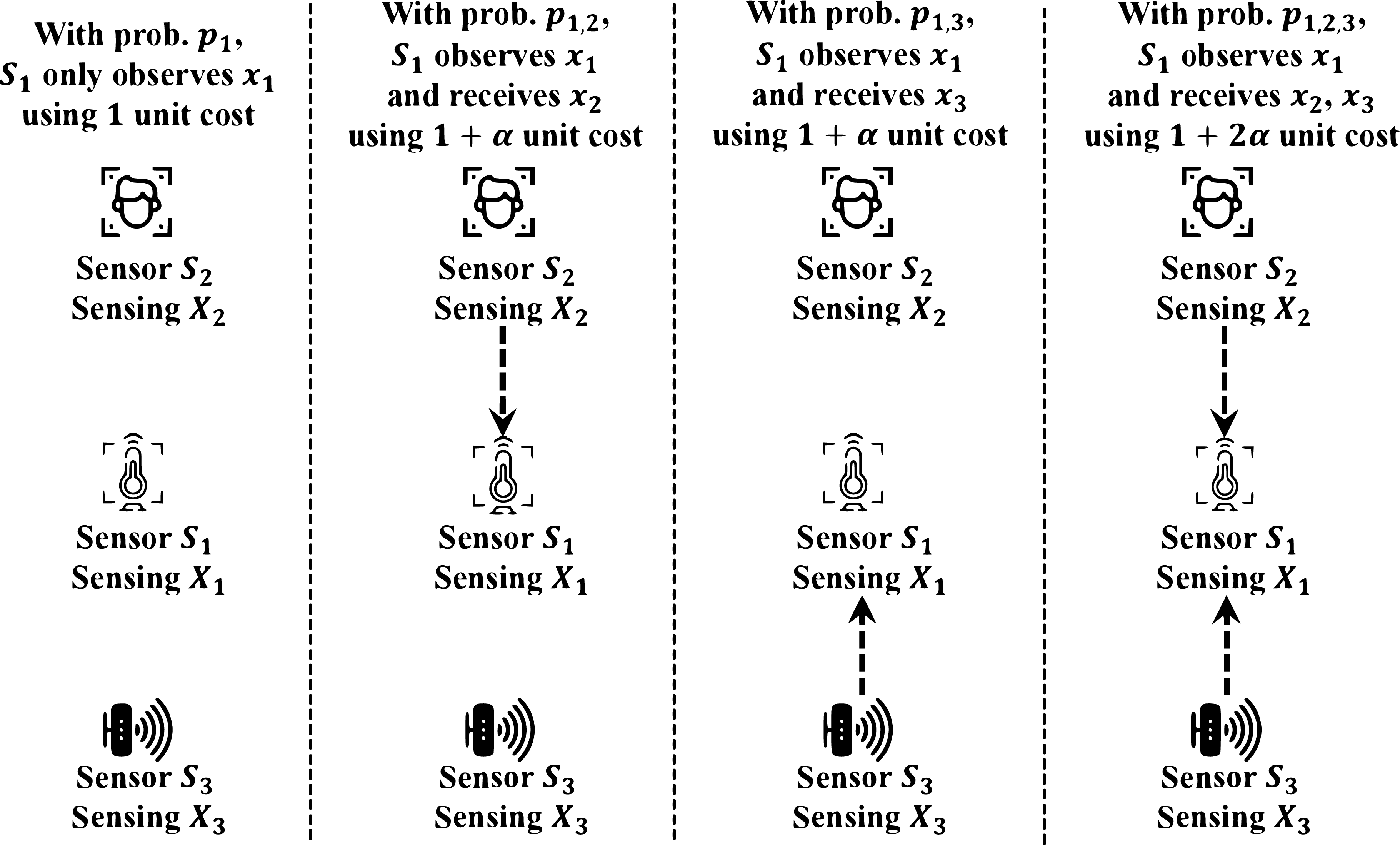}
    \begin{tabularx}{0.5\textwidth}{XXXX} \small 
    \quad (a)  & \quad \small  (b) &\quad \small  (c) & \quad \small  (d)  \\
    \end{tabularx}
    \caption{\myhl{Illustration of data collection and collaboration options from the perspective of sensor $S_1$, in a trivariate case, with {(a) no collaboration, marginal observation; (b) and (c) partial collaboration, joint bivariate observation;  (d) full collaboration, joint trivariate observation. } In \texttt{Scenario~1} and \texttt{Scenario~2}, sensors determine the sampling probabilities $p_1, p_{1,2}, p_{1,3}, p_{1,2,3}$ at the very beginning, and probabilities remain fixed throughout the considered horizon; these policies are static policies. In \texttt{Scenario 3}, sensors adaptively set the sampling probabilities $p_1(\tau), p_{1,2}(\tau), p_{1,3}(\tau), p_{1,2,3}(\tau)$ at each decision round; these policies are adaptive policies.}}\label{fig:diagram}
\end{figure}

\subsection{Distributed Parameter Estimation Problem} \label{sec:distributed}
Without loss of generality, consider sensor $S_1$. Our goal is to learn its corresponding mean parameter, $\mu_1$, under the assumption that some of the other parameters are \rev{known or unknown}. Specifically, we consider three estimation scenarios:
\begin{itemize}
	\item \texttt{Scenario 1} (only target mean unknown): sensor $S_1$ aims to estimate $\mu_1$, when the covariance matrix is known, 
    and other means, e.g., $\mu_2, ..., \mu_K$, are also known.
	\item \texttt{Scenario 2} (all means unknown): sensor $S_1$ aims to estimate $\mu_1$, when all other means, $\mu_2, ..., \mu_K$, are also unknown, while the covariance matrix is known.
    \item \texttt{Scenario 3} (correlations unknown): sensor $S_1$ aims to estimate $\mu_1$, without any knowledge of its correlations with other variables, while variances and all other means, e.g., $\mu_2, ..., \mu_K$, are known.
\end{itemize}

\rev{In the first two scenarios, we devise static sampling strategies, assuming that correlations between unknowns are given. In the third scenario, the absence of information about correlations poses additional challenges. We address these challenges through online learning, as further discussed in Section~\ref{sec:MAB-model}.}

\subsection{Data Collection and Collaboration Policies} \label{sec:datac}

 In the following, we introduce the data collection and collaboration policies considered in this work. 
 Please refer to Figure~\ref{fig:diagram} for an illustration of the actions that comprise \rev{these} policies. 

\subsubsection{Static Policy}
When the correlation coefficients $\rho_{1,k}$ between $X_1$ and other variables $X_k, \forall k \in [2,K]\equiv\{2,..,K\}$ are fixed and given (in \texttt{Scenario 1} and \texttt{Scenario 2}), we are able to derive static policies that minimize the variances of the estimators.
A static data collection policy/strategy is specified by a set of parameters $\{p_{\mathcal{K}}\in [0, 1], \forall \mathcal{K} \in \text{Pw}([K])\}$, where $\text{Pw}([K])$ denotes the power set of $[K]\equiv\{1,2,...,K\}$. $p_{\mathcal{K}}\in [0, 1]$ denotes the probability that \textit{only} sensors $S_k$, $k \in \mathcal{K}\subset[K]$ are active and \textit{only} variables $X_k$, $k\in\mathcal{K}$ are observed at each time. For example, $p_\emptyset$ denotes the probability that no observation is made; $p_1$ (resp., $p_2,...,p_K$) denotes the probability that \textit{only} a marginal observation $X_1$ (resp., $X_1,..., X_K$) is made; and $p_{1,2}$ (resp., $p_{1,3},..., p_{K-1,K}$) denotes the probability that \textit{only} a joint observation $(x_1, x_2)$ (resp., $(x_1, x_3), ..., (x_{K-1}, x_K)$) is made.  Note that, by definition, $\sum_{\mathcal{K}} p_{\mathcal{K}} = 1$. Similarly, we denote the number of each type of observations we obtain as $\{N_{\mathcal{K}}\in [0, 1], \forall \mathcal{K} \in \text{Pw}([K])\}$, e.g., $N_1$ is the number of $x_1$ samples we have and $N_{1,2}$ is the number of $(x_1, x_2)$ samples we have. Under the static data collection strategy, the expected number of each type of sample we obtain up to time $t$ is determined by the sampling probability parameters, e.g., $\mathbb{E}[N_1] = p_1 \cdot t$ and $\mathbb{E}[N_{1,2}] = p_{1,2} \cdot t$.

\subsubsection{Multi-Armed Bandit-Based Adaptive Policy}\label{sec:MAB-model}  
When the correlation coefficients $\rho_{1,k}$ are unknown (in \texttt{Scenario 3}), we face a ``chicken and egg'' problem where estimating correlations is key to determining the optimal sampling strategy, but the estimation of correlation itself requires sampling.  Multi-armed bandits are a natural tool to address such a conundrum, as they provide a method to balance collecting additional samples (exploration) and using the available data optimally (exploitation). 

We further model our data collection and collaboration problem as a MAB  problem~\cite{bubeck2012regret}.
In the MAB terminology, each available action is called an \emph{arm}.  In our problem, each arm corresponds to a subset of variables to be sampled.    
To prevent the number of arms (the exploration space) from growing combinatorially large, we restrict ourselves to only consider collecting either univariate marginal observations or bivariate joint observations. 
Hence, our multi-armed bandit model has arms $j \in [K]$, where arm $1$ corresponds to collecting the sample(s) of marginal observation $X_1$ and arm $j \in [2, K]$ corresponds to collecting the sample(s) of bivariate joint observation $(X_1, X_j)$.   \myhl{Figure~\ref{fig:diagram} illustrates a trivariate scenario where arms 1, 2 and 3 correspond to Figure~\ref{fig:diagram}(a),~\ref{fig:diagram}(b) and~\ref{fig:diagram}(c), respectively. }

\begin{table}[t]
\caption{Multi-armed Bandits Inter-Decision Time -- more stringent resource constraints yield longer inter-decision times. }
\centering
\begin{threeparttable}[t]
\scriptsize
\begin{tabular}{l|l|l|l}
\toprule
Case &  Level of & Time slots per & Number of  \\
& Constraint &  decision round & marginal samples\tnote{*}  \\
\midrule
$E \geq \alpha + 1$ & Unconstrained & 1 & 1\\
\midrule
$1 \leq E < \alpha + 1$ &  Mild  & 
 $\lceil(\alpha+1)/E\rceil$ & $\lceil(\alpha + 1)/E\rceil$   \\
\midrule
$E < 1$ & Stringent  & $\lceil(\alpha+1)/E\rceil$ & $\lfloor\alpha + 1\rfloor$  \\
\bottomrule
\end{tabular} 
\begin{tablenotes}
     \item[*] At each decision round, the multi-armed bandit-based policy can collect a bivariate observation $(X_1, X_j)$, where $j \in [2, K]$, or the number of marginal (univariate) samples  $X_1$ given by the last column of the table
   \end{tablenotes}
\end{threeparttable}
\label{tab:mabresconst}
\end{table}

At each \textit{decision round} $\tau$, a multi-armed bandit learner selects one arm, $J_{\tau} \in [K]$, to pull, i.e., it makes one decision. 
To fulfill resource constraint requirements, we only allow the multi-armed bandit learner to make one decision every several time slots.  Stringent constraints yield longer inter-decision times, as detailed in Table~\ref{tab:mabresconst}.  We denote the total number of decision rounds by $\mathcal{T}$, where $\mathcal{T} \leq T$.

In an MAB-based adaptive policy, the probability of selecting each of the arms can vary over time.   For this reason, in the MAB setting, we make the dependence of $p$ on $\tau$ explicit.  
The control variables of the MAB-based  policy are given by 
$\{p_1{(\tau)}, p_{1, j}{(\tau)} \in \{0,1\}, \forall j\in[2, K], \forall {\tau} \in [\mathcal{T}]\}$. 
For each decision round $\tau$, it is required that $p_1{(\tau)} + \sum_{j=2}^K p_{1,j}{(\tau)} = 1$. For example, here $p_1{(\tau)} = 1$ (resp. $p_{1, 2}{(\tau)} = 1$, $p_{1, 3}{(\tau)} = 1$, ..., or $p_{1, K}{(\tau)} = 1$) denotes that arm $1$ (resp. arm $2, 3, ..., K$) is selected and sample(s) of marginal observation $X_1$ (resp. bivariate observation $(X_1, X_2), (X_1, X_3), ..., (X_1, X_K)$) is/are collected at decision round $\tau$.


\section{\myhl{\texttt{Scenario 1} - Correlation Information Available and Only Target Mean Unknown}} \label{sec:correlation-known-1}
We begin by studying \texttt{Scenario 1}, 
where the covariance matrix is known, and all mean parameters other than our target mean parameter are also known. This scenario models the case that a sensor/agent is newly allocated into a sensor network or a multi-agent learning group while other sensors have learned their corresponding parameters, and we know the covariance matrix, e.g., due to geographic relationships.

In the following, we first analyze bivariate/two-sensor and/or trivariate/three-sensor case(s) to provide intuition and then discuss the general multivariate (multiple-sensor) case. 
We formally formulate our estimation problem as an expected constrained Fisher information (FI)~\cite{cover1999elements} optimization problem, analyze the FI to resource cost ratio of different types (marginal/joint) of samples, and derive closed-form optimal static sample collection/resource allocation policies and estimators given various correlation values and resource budgets. 

We show that, in \texttt{Scenario 1}, the correlation structure among sensors can be leveraged to estimate the parameter of interest collaboratively better, i.e., a higher rate of convergence, than estimating it individually.

\subsection{Bivariate (Two-sensor) Case}\label{sec:scenario-1-bivariate}
As sensor $S_1$ aims to estimate $\mu_1$, its objective is to maximize the FI regarding $\mu_1$, $\mathcal{I}(\mu_1)$. 
The constrained expected FI maximization problem can be formally written as:
\begin{align}
&\max\limits_{p_{\emptyset}, p_1, p_2, p_{1,2}\in [0, 1]}\,\,  p_1 \mathcal{I}_{X_1}(\mu_1) + p_{1,2} \mathcal{I}_{(X_1,X_2)}(\mu_1) \label{eq:decentalized:one-unknown:rho-known1}\\
&\qquad\qquad\qquad\quad\,\,\,\,\,= p_1\frac{1}{\sigma_1^2} +p_{1,2}\frac{ 1 }{(1-\rho_{1,2}^2)\sigma_1^2}, \\
&\,\,\text{s.t. }\,\,  p_{\emptyset} + p_1 + p_2 + p_{1,2} = 1,\\
&\quad\quad\, p_1 + (\alpha +1) p_{1,2} \leq E.
\label{eq:decentalized:one-unknown:rho-known2}
\end{align}
Note that $p_2$ does not appear in the objective function of this optimization problem because marginal observations from $X_2$, not paired up with observations of $X_1$ to produce joint observations, add no information about $\mu_1$, i.e., $\mathcal{I}_{X_2}(\mu_1) = 0$.

\begin{figure}[t]
    \centering
        \includegraphics[width=0.468\linewidth]{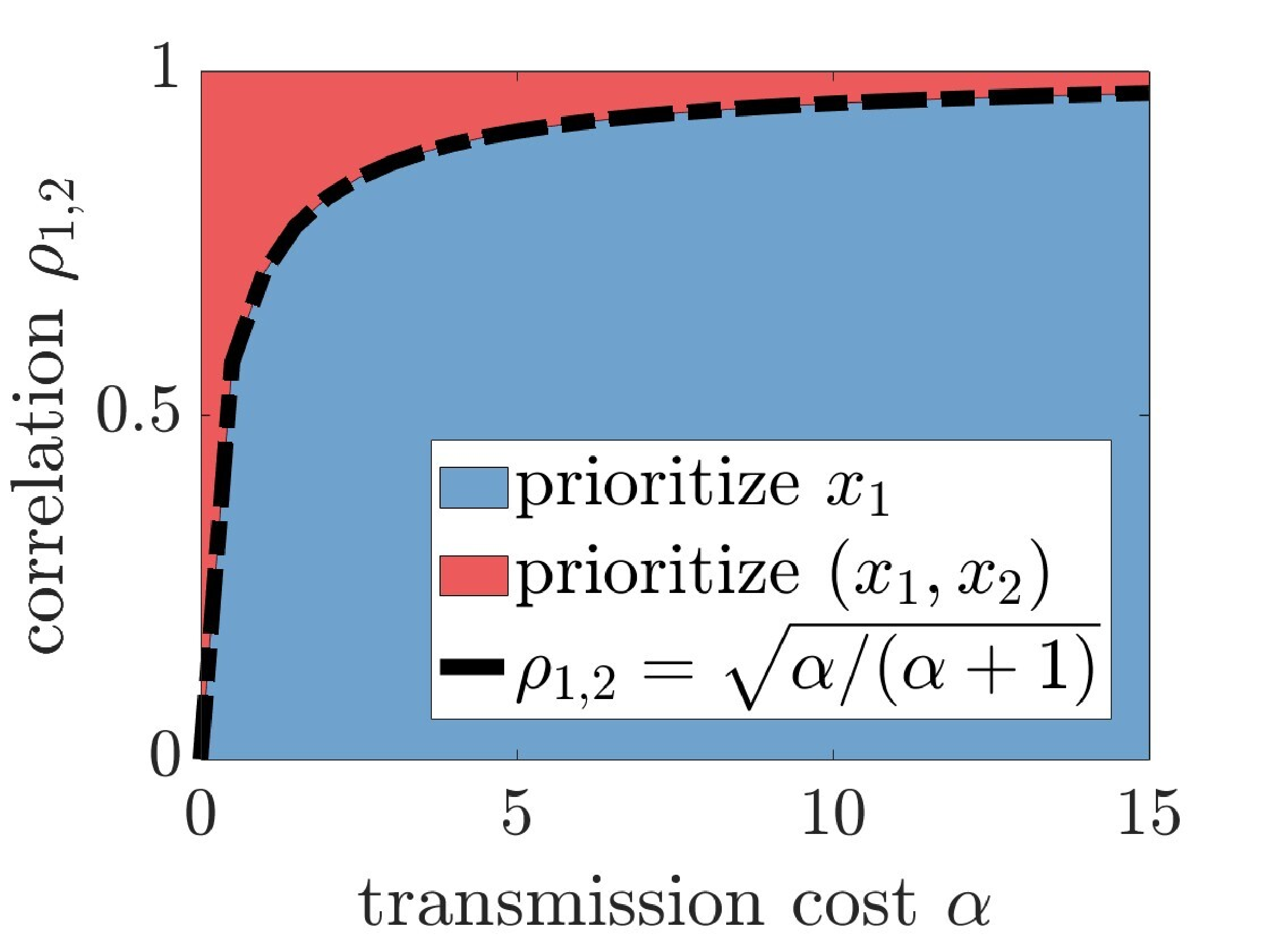}
        \caption{\texttt{Scenario 1} Bivariate (Two-Sensor) Case: critical threshold of correlation for prioritizing collecting joint observations over collecting marginal observations under various transmission resource costs $\alpha$}
        \label{fig:bivariate_critical}
\end{figure}

\myhl{Next we determine}  whether to \textit{prioritize} samples containing only information about $X_1$ or joint samples about $(X_1, X_2)$.  In what follows, we show that the correlation coefficient plays an important role in that decision. Intuitively, the benefits from collecting joint samples \rev{should} increase with correlation.  
A joint sample, $(x_1, x_2)$, contains 
$\mathcal{I}_{(X_1,X_2)}(\mu_1) = 1/((1-\rho_{1,2}^2)\sigma_1^2)$
amount of Fisher information about parameter $\mu_1$ and costs $\alpha+1$ units of resources. The resources used to collect $(x_1, x_2)$ can alternatively be used to collect $\alpha+1$ samples from  $X_1$, which in total contain 
$(\alpha+1)\mathcal{I}_{X_1}(\mu_1) = (\alpha+1)/\sigma_1^2$
amount of Fisher information about parameter $\mu_1$. Hence, when 
\begin{equation}
{1}/({(1-\rho_{1,2}^2)\sigma_1^2})>(\alpha+1)/{\sigma_1^2} \quad\text{or}\quad \rho_{1,2}^2>{\alpha}/({\alpha+1}), \label{eq:condikey}
\end{equation}
a joint observation, $(x_1, x_2)$, provides more information than $\alpha+1$ marginal observations solely from $S_1$.  The lower the resource cost $\alpha$, the lower the minimum value of $\rho_{1,2}$ that motivates collecting joint observations.   In particular, when there is no transmission cost, i.e., $\alpha=0$, 
we should set $p_{1,2}=1$ as $\mathcal{I}_{(X_1,X_2)}(\mu_1) \geq \mathcal{I}_{X_1}(\mu_1)$.   Figure~\ref{fig:bivariate_critical} illustrates the regions where collecting marginal observations only from $X_1$ or joint observations of $(X_1, X_2)$ should be prioritized. The dashed line separating the two regions corresponds to  $\rho_{1,2}=\sqrt{\alpha/(\alpha+1)}$.  

\begin{figure}[t]
    \vskip -0.1in
    \centering
    \subfloat{\includegraphics[width=0.5\linewidth]{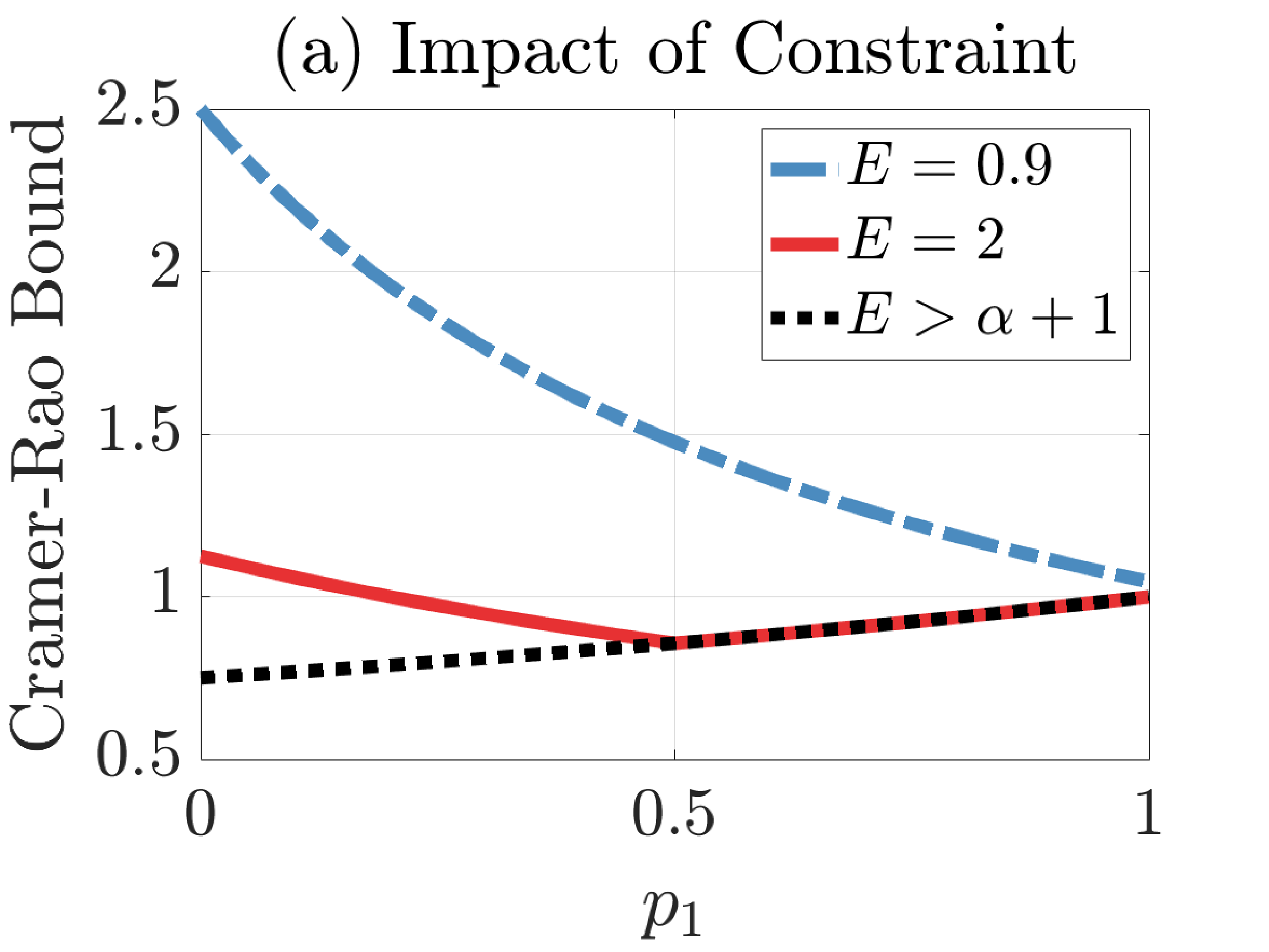}\label{subfig:decentralized:known-rho-CRB}}
    \hfill
    \subfloat{\includegraphics[width=0.5\linewidth]{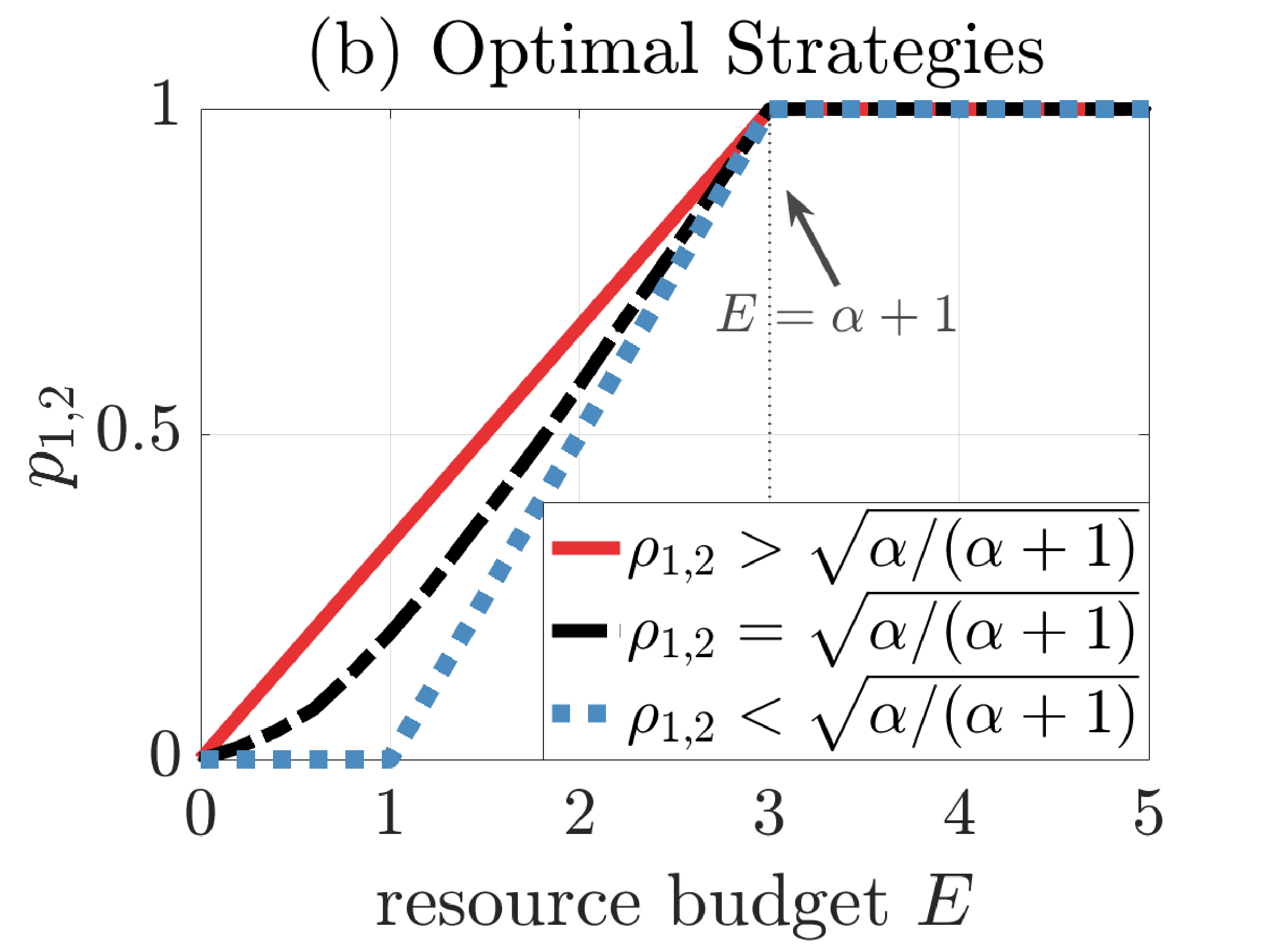}\label{subfig:decentralized:known-rho-optimal-strategies}}
    \caption{\texttt{Scenario 1} Bivariate (Two-Sensor) Case: (a) (expected) Cramer-Rao bound under various static data collection policies (various values of $p_1$ and correspondingly $p_{1, 2} = \min\{1-p_{1}, (E-p_1)/(\alpha+1)\}$) when transmission cost $\alpha=2$, correlation $\rho_{1,2}=0.5$, and variance $\sigma_1^2 = \sigma_2^2 =1$; (b) Optimal data collection strategies (denoted by $p_{1,2}$, with corresponding $p_1 = \min\{1-p_{1,2}, E-(1+\alpha)p_{1,2}\}$) under different correlations and resource budgets}
\end{figure}

Given the above prioritization scheme, we now consider how to allocate limited resources to sampling each type of observation in the constrained optimization problem,~\eqref{eq:decentalized:one-unknown:rho-known1}-\eqref{eq:decentalized:one-unknown:rho-known2}.
Figure \ref{subfig:decentralized:known-rho-CRB} illustrates how constraint~\eqref{eq:decentalized:one-unknown:rho-known2} affects the optimization problem.
In Figure \ref{subfig:decentralized:known-rho-CRB}, we fix transmission cost $\alpha=2$, correlation $\rho_{1,2}=0.5$, variance $\sigma_1^2 = \sigma_2^2 =1$, and plot the corresponding Cramer-Rao bound (i.e., the reciprocal of the FI,~\eqref{eq:decentalized:one-unknown:rho-known1}) under various resource constraints and sampling strategies, i.e., $p_1$ and $p_{1, 2} = \min\{1-p_{1}, (E-p_1)/(\alpha+1)\}$.
\rev{When there are no resource constraints,} e.g., $E = \infty > \alpha+1$ (black dotted line), the Cramer-Rao bound (CRB) is minimized when $p_1=0$ and $ p_{1, 2}=1$.
That is, one always prefers collaborating to collect joint observations.
When accounting for a resource budget of $E=2$ (red solid line), the resource constraint \eqref{eq:decentalized:one-unknown:rho-known2} is inactive in the the region where $p_1 \geq 0.5$ and the value of $p_{1,2}$ can be set to $1-p_{1}$; the resource constraint \eqref{eq:decentalized:one-unknown:rho-known2} becomes active in the region where $p_1 < 0.5$, and hence, the value of $p_{1,2}$ is determined by the resource constraint: $p_{1,2} = (E-p_1)/(\alpha+1) \leq (1-p_{1})$. As the CRB is minimized at $p_1=p_{1,2}=0.5$, spending a portion of the budget on local data collection and the remainder of the budget on collaboration is preferable to investing the entire budget on marginal samples or on joint samples exclusively.
Under more stringent resource constraints (blue dashed line), data sharing becomes prohibitive, and the CRB is minimized when $p_1=1$.

The optimal data collection policies/strategies under different resource budgets are summarized in Table~\ref{tab:optimal_policy} and illustrated in Figure~\ref{subfig:decentralized:known-rho-optimal-strategies}. 
\begin{itemize}
    \item\rev{When $\rho_{1,2}^2  < \alpha/(\alpha+1)$, the optimal strategy is to make marginal observations on $X_1$ (denoted by the blue dotted line). Any residual resource budget is used to make joint observations.} 
    \item\rev{When $\rho_{1,2}^2 > \alpha/(\alpha+1)$, the entire resource budget should be used to collect joint observations  (denoted by the red solid line). }
    \item\rev{When $\rho_{1,2}^2 = {\alpha}/({\alpha+1})$, any choices of $p_1$ and $p_{1,2}$ that lie between the red solid line and the blue dotted line in Figure \ref{subfig:decentralized:known-rho-optimal-strategies}, such as the black dash-dot line, are optimal. In this case, the two strategies mentioned above are also optimal.}
\end{itemize}

\begin{table}[t]
    \caption{\texttt{Scenario 1} Bivariate (Two-Sensor) Case: Optimal Static Data Collection Policy under Different Correlations and Resource Budgets}\label{tab:optimal_policy}
    \centering
    \begin{tabular}{ c|c|c|c } 
     \toprule
      & $E\leq1$ & $1 < E < \alpha+1$& $E\geq \alpha+1$ \\ \midrule
      $\rho_{1,2}^2 \geq \frac{\alpha}{\alpha+1}$ & $p_{1,2} = \frac{E}{\alpha+1}$ & $p_{1,2} = \frac{E}{\alpha+1}$ & $p_{1,2} = 1$\\ 
     &$p_1 = 0$&$p_1 = 0$&$p_1 = 0$\\\midrule
      $\rho_{1,2}^2 \leq \frac{\alpha}{\alpha+1}$ & $p_{1,2} = 0$ & $p_{1,2} = \frac{E-1}{\alpha}$ & $p_{1,2} = 1$\\ 
     &$p_1 = E$&$p_1=\frac{\alpha+1-E}{\alpha}$&$p_1 = 0$\\ 
     \bottomrule
    \end{tabular}
\end{table}

Once the data is collected, we estimate the parameter of interest, $\mu_1$, using Kalman filtering~\cite{kalman1960new, pei2019elementary}. 
Specifically, with $N_{1}$ i.i.d. marginal samples, $x_{1,1},...,x_{1,{N_1}}$, we have a straightforward unbiased estimator, namely the sample mean,
\begin{equation}\label{eq:delta_1}
    \delta_1 \equiv \bar{x}_1 \equiv \frac{1}{N_1}\sum_{i=1}^{N_1} x_{1,i},
\end{equation}
with variance $\text{Var}(\delta_1) = \sigma_1^2/N_1 = 1/(N_1\mathcal{I}_{X_1}(\mu_1))$.
With $N_{1,2}$ i.i.d. joint (two-sensor) samples, 
we have the uniformly minimum-variance unbiased estimator (UMVUE)~\cite{casella2002statistical, bishwal2008note},
\begin{align}\label{eq:delta_12}
    &\delta_{1,2} \equiv \bar{x}_1 - \beta(\bar{x}_2 - \mu_2), &\beta \equiv\rho_{1,2}\sigma_1/\sigma_2,
\end{align}
whose variance is $\text{Var}(\myhl{\delta_{1,2}}) = \sigma_1^2(1-\rho_{1,2}^2)/N_{1,2} = 1/(N_{1,2}\mathcal{I}_{(X_1, X_2)}(\mu_1))$.
Kalman filtering fuses the estimates we have through a linear combination, i.e., 
\begin{equation}\label{eq:kalman-uni-bi}
    \delta^* = g_1 \delta_1 + g_{1,2} \delta_{1,2}, 
\end{equation}
where the weights $g_1, g_{1,2} \in [0, 1],$ and $g_1 + g_{1,2} = 1$.
Here, the Kalman filtering estimator \rev{(KFE)} $\delta^*$ has variance $\text{Var}(\delta^*) = g_1^2 /(N_1\mathcal{I}_{X_1}(\mu_1)) + g_{1,2}^2 /(N_{1,2}\mathcal{I}_{(X_1, X_2)}(\mu_1))$. Note that $\text{Var}(\delta^*)$ is at most as large as $\max\{\text{Var}(\delta_1), \text{Var}(\myhl{\delta_{1,2}})\}$, and with proper choice of weights $g_1, g_{1,2}$, $\text{Var}(\delta^*)$ can even be smaller than both $\text{Var}(\delta_1)$ and $\text{Var}(\myhl{\delta_{1,2}})$. Intuitively, the estimate in which we have more confidence, i.e., the estimate with smaller variance, should be given a larger weight. In fact, $\text{Var}(\delta^*)$ is minimized when 
\begin{align}
    g_1 &
    = \frac{N_1 (1-\rho_{1,2}^2)}{N_1(1-\rho_{1,2}^2)+N_{1,2}},\label{eq:g_1}\\
    g_{1,2} &
    = \frac{N_{1,2}}{N_1(1-\rho_{1,2}^2)+N_{1,2}}.\label{eq:g_12}
\end{align}
As the value of $\rho_{1,2}$ is available in \texttt{Scenario 1}, we can easily compute this set of optimal weights $g_1, g_{1,2}$ according to~\eqref{eq:g_1} and~\eqref{eq:g_12} and use it in the  \rev{KFE}. Finally, it is worth noting that the \rev{KFE} with optimal $g_1, g_{1,2}$ is also the maximum likelihood estimator, \rev{as we show} in Appendix~\ref{sec:likelihood-scenario1}. \rev{In other words, the KFE with optimal weights  effectively utilizes the samples collected by the optimal data collection strategy to achieve the Cramer-Rao bound. }

\subsubsection{Trivariate (Three-sensor) Case}
In this case, sensor $S_1$ can choose to observe $X_1$ solely or also collect observation(s) from one/both of the other sensors $S_2$, $S_3$ at each time slot (see \myhl{Figure~\ref{fig:diagram}}). Hence, the constrained FI optimization problem becomes
\begin{align}
&\max\limits_{p_{\mathcal{K}} \in [0,1], \mathcal{K} \in \text{Pw}([3]) } \,\, p_1 \mathcal{I}_{X_1}(\mu_1) + p_{1,2}\mathcal{I}_{(X_1, X_2)}(\mu_1) +\notag\\ 
&\qquad\qquad\quad p_{1,3} \mathcal{I}_{(X_1, X_3)}(\mu_1)+ p_{1,2,3}  \mathcal{I}_{(X_1, X_2, X_3)}(\mu_1), \\
&\,\,\text{s.t.}\,\,  \sum\limits_{\mathcal{K} \in \text{Pw}([3])}p_\mathcal{K} = 1,\\
&\quad\,\,\,\,\,\, p_1 + (1+\alpha)p_{1,2} + (1+\alpha)p_{1,3} + (1+2\alpha)p_{1,2,3} \leq E. 
\end{align}
Note that $p_2$, $p_3$, and $p_{2,3}$ are set to zero because $I_{X_2}(\mu_1)$, $I_{X_3}(\mu_1)$, and $I_{(X_2, X_3)}(\mu_1)$ are zero.

\begin{figure*}[t]
   \centering
        \subfloat{\includegraphics[width=0.26\linewidth]{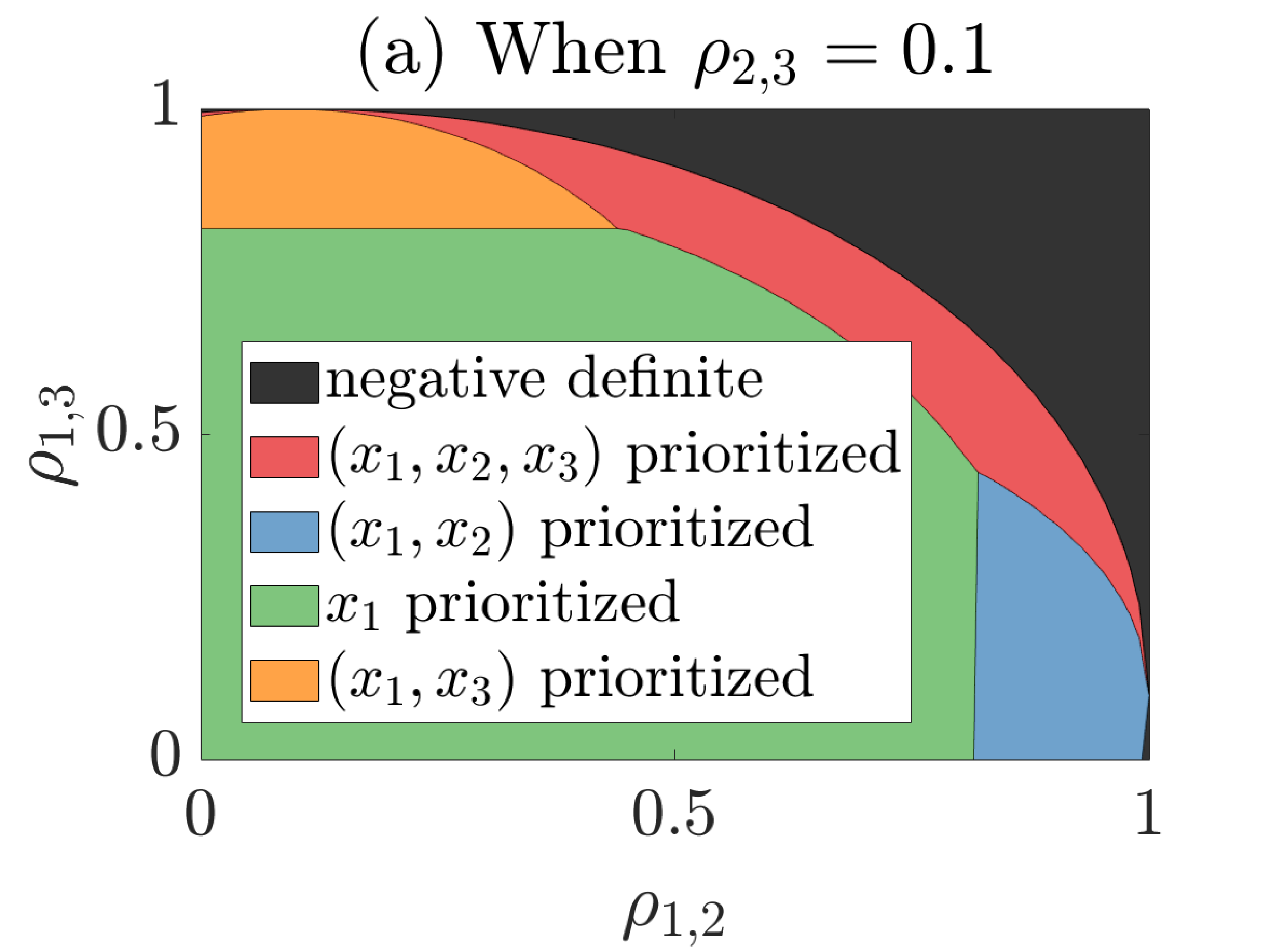}\label{subfig:trivariate-rho01}}
        \hfil
        \subfloat{\includegraphics[width=0.26\linewidth]{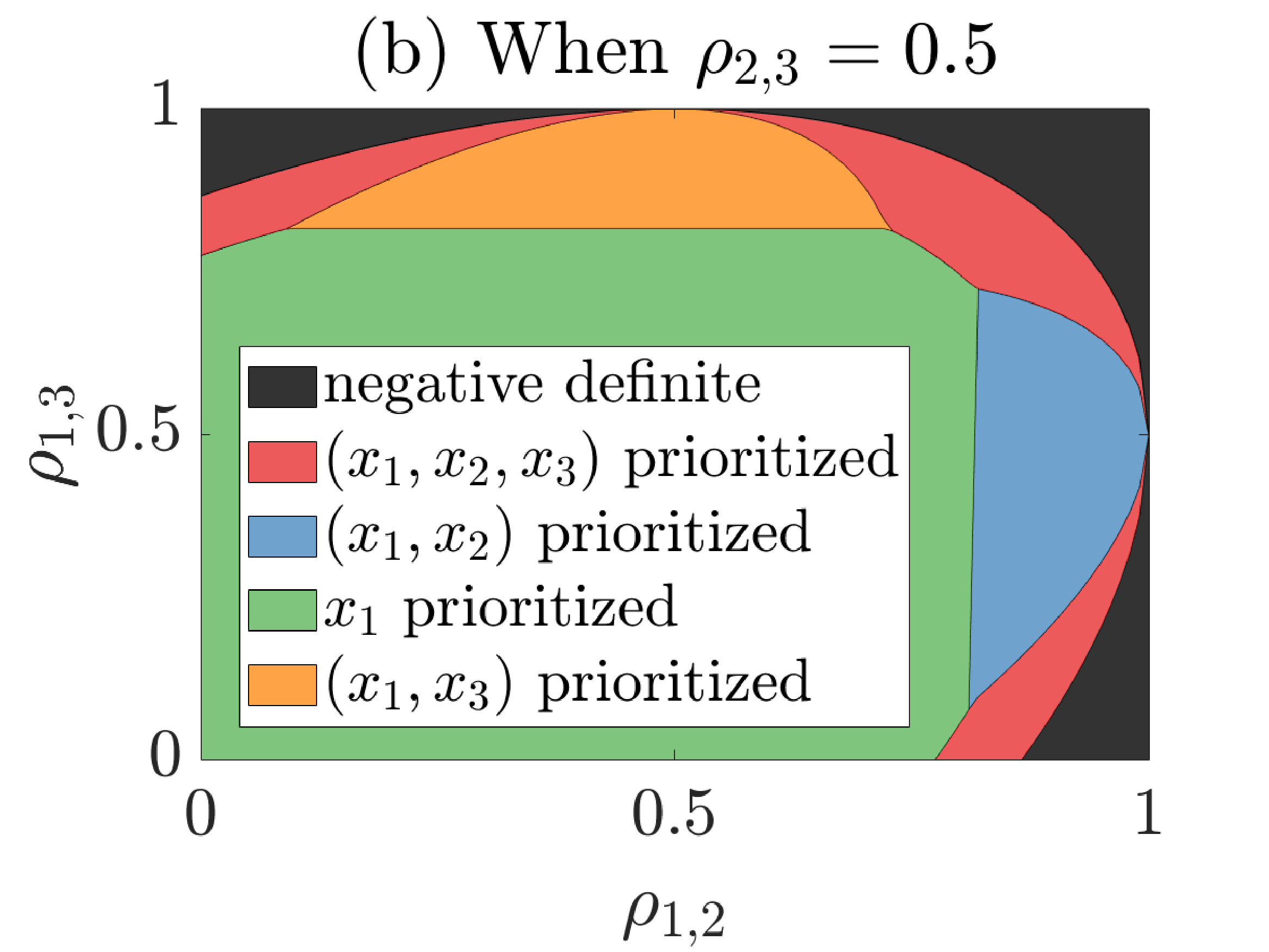}\label{subfig:trivariate-rho05}}
        \hfil
        \subfloat{\includegraphics[width=0.26\linewidth]{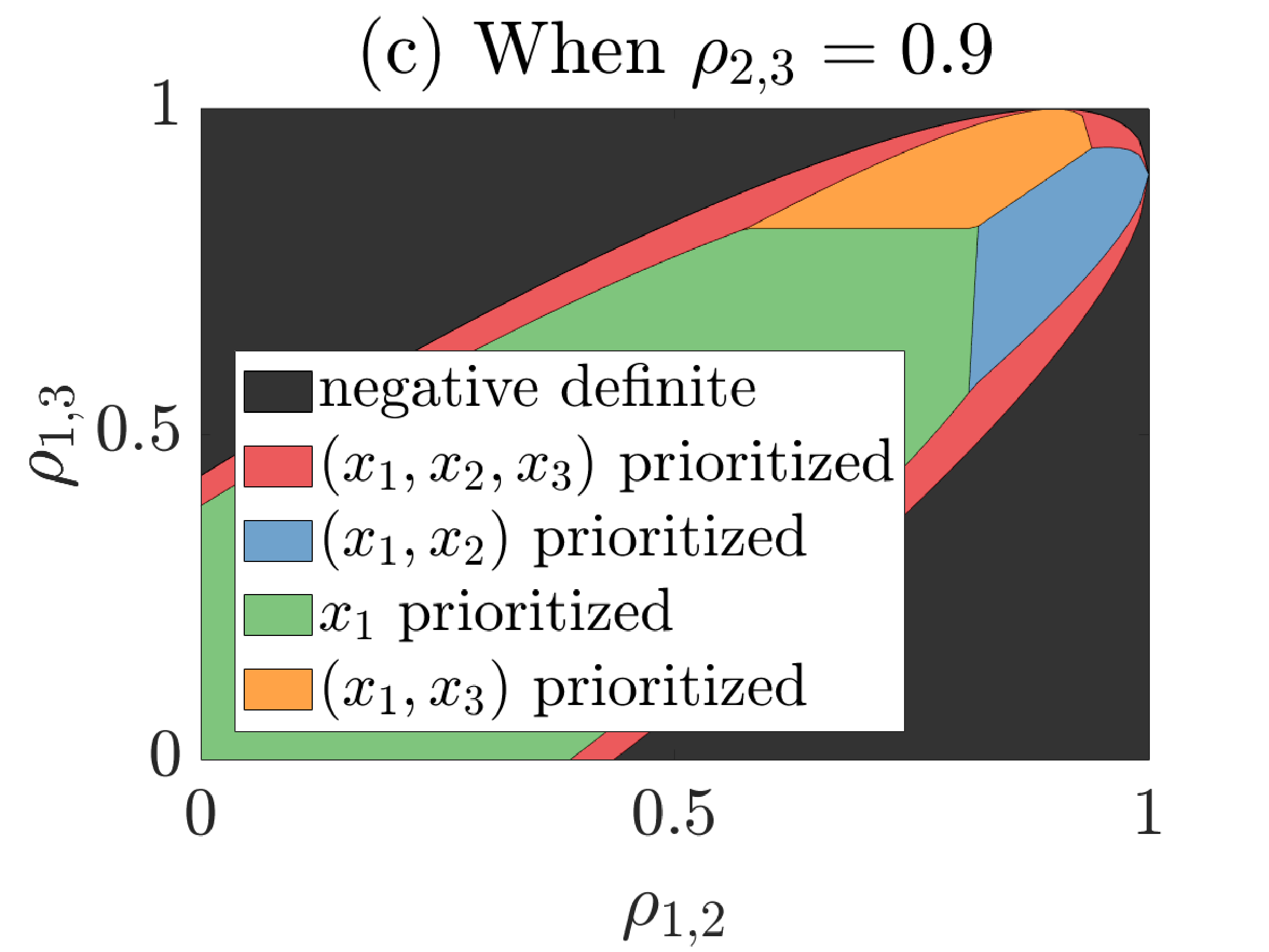}\label{subfig:trivariate-rho09}}
        \caption{\texttt{Scenario 1} Trivariate (Three-Sensor) Case: critical thresholds of correlations $\rho_{1,2}$ and $\rho_{1,3}$ for prioritizing different types (univariate, bivariate, or trivariate) of samples when $\rho_{2,3}$ and transmission resource cost $\alpha = 2$ are fixed}\label{fig:trivariate}
\end{figure*}

In the following, we analyze the prioritization of collecting samples of each type (univariate, bivariate, or trivariate). From the bivariate case above, we learn that a bivariate joint sample of $(X_1, X_2)$ (resp. $(X_1, X_3)$), provides more information than $\alpha+1$ univariate samples of $X_1$ when $\rho_{1,2}^2 > \alpha/(\alpha+1)$ (resp. $\rho_{1,3}^2 > \alpha/(\alpha+1)$). \myhl{In addition, between the two types of bivariate}  samples, a sample of $(X_1, X_2)$ contains more information than a sample of $(X_1, X_3)$ when $\rho_{1,2} > \rho_{1,3}$. It remains to determine when trivariate samples should be prioritized over bivariate or univariate samples. The FI of a  trivariate joint sample of $(X_1, X_2, X_3)$ regarding $\mu_1$ is 
\begin{align}
    &\mathcal{I}_{(X_1, X_2, X_3)}(\mu_1) = \left(\mathbf{\Sigma}_{(X_1, X_2, X_3)}^{-1}\right)^{1, 1}\\
    &= \frac{1-\rho_{2,3}^2}{(1 + 2\rho_{1,2}\rho_{1,3}\rho_{2,3} - \rho_{1,2}^2 - \rho_{1,3}^2 - \rho_{2,3}^2)\sigma_1^2}.
\end{align}
The resource needed to collect a trivariate sample can alternatively be used to collect $(2\alpha+1)/(\alpha+1)$ bivariate samples or $2\alpha+1$ univariate samples. Considering the Fisher information of each type of sample, a trivariate sample $(x_1, x_2, x_3)$ provides more information than $(2\alpha+1)/(\alpha+1)$ bivariate samples of $(X_1, X_2)$ when
\begin{equation}\label{eq:trivariate-prior-bi}
    \alpha < \frac{\rho_{1,3}^2+\rho_{1,2}^2\rho_{2,3}^2-2\rho_{1,2}\rho_{1,3}\rho_{2,3}}{1+4\rho_{1,2}\rho_{1,3}\rho_{2,3}-\rho_{1,2}^2-\rho_{2,3}^2-\rho_{1,2}^2\rho_{2,3}^2-2\rho_{1,3}^2}.
\end{equation}
Similarly, by replacing $\rho_{1,2}$ (resp. $\rho_{1,3}$) with $\rho_{1,3}$ (resp. $\rho_{1,2}$) in~\eqref{eq:trivariate-prior-bi}, we obtain the condition under which a trivariate joint sample,$(x_1, x_2, x_3)$ provides more information than $(2\alpha+1)/(\alpha+1)$ bivariate samples of $(X_1, X_3)$. Besides, a trivariate joint sample, $(x_1, x_2, x_3)$, provides more information than $2\alpha+1$ univariate samples of $X_1$ when
\begin{equation}
    \alpha < \frac{\rho_{1,2}^2+\rho_{1,3}^2-2\rho_{1,2}\rho_{1,3}\rho_{2,3}}{2(1+2\rho_{1,2}\rho_{1,3}\rho_{2,3}-\rho_{1,2}^2-\rho_{1,3}^2-\rho_{2,3}^2)}.
\end{equation}
The prioritization rules still depend on the correlation coefficients and are easily calculated; however, the rules are now more complicated, as illustrated in Figure~\ref{fig:trivariate}. For example, in the blue area in Figure~\ref{subfig:trivariate-rho05}, collecting sample $(x_1, x_2)$ is prioritized as $\rho_{1,2}$ is large while $\rho_{1,3}$ is neither too large nor too small. Interestingly, if $\rho_{1,3}$ were smaller, the trivariate sample $(x_1, x_2, x_3)$ should be prioritized, which is not intuitive given the intuition we learn in the bivariate (two-sensor) case in the previous subsection.  \myhl{While in the bivariate case, an additional variable is beneficial only when the correlation is large enough (see, e.g., Figure~\ref{fig:bivariate_critical}), in the trivariate case, an additional variable might become beneficial if its correlation is reduced.} 

Computing data collection and collaboration strategies under resource constraints, fortunately, is still feasible despite the increasing set of prioritization rules. This is because the constrained expected Fisher information optimization problem in \texttt{Scenario 1} is a linear programming problem. \myhl{In particular, note that as the FI of each sample type can be calculated in advance, the coefficients of the linear optimization function are fixed and given.}

\myhl{Next, we derive optimal estimators for the three-sensor case, leveraging the four  types of samples illustrated in Figure~\ref{fig:diagram}.}  
 With univariate or bivariate samples, estimators~\eqref{eq:delta_1} and~\eqref{eq:delta_12} can be applied respectively to produce an estimate of $\mu_1$. As for trivariate samples, by the factorization theorem and the Lehmann-Scheffe theorem~\cite{casella2002statistical, bishwal2008note}, the UMVUE is given by
\begin{align}
    \delta_{1,2,3} = \bar{x}_1 &- \frac{(\rho_{1,2}-\rho_{1,3}\rho_{2,3})\sigma_1}{(1-\rho_{2,3}^2)\sigma_2}(\bar{x}_2-\mu_2) - \nonumber \\
    &- \frac{(\rho_{1,3} - \rho_{1,2}\rho_{2,3})\sigma_1}{(1-\rho_{2,3}^2)\sigma_3}(\bar{x}_3-\mu_3).
\end{align}
Using the additional types of samples in the trivariate case, the Kalman filtering estimator \rev{(KFE)} can be extended to
\begin{align}
    &\delta^* = g_1 \delta_1 + g_{1,2} \delta_{1,2} + g_{1,3}\delta_{1,3} + g_{1,2,3} \delta_{1,2,3},\\
    &g_1 +g_{1,2} + g_{1,3} + g_{1,2,3} = 1,
\end{align}
where $\delta_{1,3}$ has the same form as $\delta_{1,2}$ in~\eqref{eq:delta_12}, and optimal weights $g_1$, $g_{1,2}$, $g_{1,3}$, $g_{1,2,3}$ can be computed given the number of samples of each type, \myhl{given by}  $N_1$, $N_{1,2}$, $N_{1, 3}$, $N_{1,2,3}$.

\subsubsection{Multivariate (Multiple-sensor) Case}
With insights gained from analyzing the bivariate and trivariate cases, we observe that, in \texttt{Scenario 1}, as the Fisher information of each type of sample can be computed a priori, the constrained expected Fisher information optimization problem is  a linear programming problem: 
\begin{align}
&\max\limits_{p_{\mathcal{K}}, \mathcal{K} \in \text{Pw}([K])} \,\, p_1 \mathcal{I}_{X_1}(\mu_1)+ \sum_{k\in[2, K]} p_{1,k}\mathcal{I}_{(X_1,X_k)}(\mu_1)\notag\\ 
&\qquad\qquad\qquad+\sum_{k_1, k_2\in[2, K]} p_{1, k_1, k_2}\mathcal{I}_{(X_1, X_{k_1}, X_{k_2})}(\mu_1)\notag\\
&\qquad\qquad\qquad+...+p_{1,2,...,K}\mathcal{I}_{(X_1, X_2,..., X_K)}(\mu_1), \label{eq:scenario-1-multivariate}\\
&\text{s.t. }\,\,  \sum_{\mathcal{K}\in \text{Pw}([K])} p_{\mathcal{K}} \leq 1,\\
&\qquad p_1 + \sum_{k\in[2, K]} (\alpha+1) p_{1,k}\notag+ \sum_{k_1, k_2\in[2, K]} (2\alpha+1)p_{1, k_1, k_2}\notag\\
&\qquad\,\,\,\,\,\, +...+((K-1)\alpha+1)p_{1,2,...,K} \leq E.
\end{align}
Given the correlation information $\rho_{1,2}, \rho_{1,3}, ..., \rho_{K-1, K}$, transmission cost $\alpha$, and resource budget $E$, the optimal static data collection policy $p_{\mathcal{K}}, \mathcal{K} \in \text{Pw}([K])$ can be computed and used to control the sampling process.
To take advantage of the different types of samples collected, we can fuse the samples using a Kalman filter,
\begin{align}
    &\delta^* = \sum_{\mathcal{K}\in \text{Pw}([K])} g_{\mathcal{K}} \delta_{\mathcal{K}}, &\sum_{\mathcal{K}\in \text{Pw}([K])} g_{\mathcal{K}}= 1,
\end{align}
where $\delta_{\mathcal{K}}, \mathcal{K}\in \text{Pw}([K])$ are the UMVUEs of $\mu_1$.


\section{\myhl{\texttt{Scenario 2} - Correlation Information Available and All Means Unknown}}  \label{sec:correlation-known-2}

In the following, we analyze \texttt{Scenario 2}
This scenario corresponds to a setting where the covariance matrix is known, but all mean parameters are unknown. 
None of the sensors/agents have learned their corresponding mean parameters yet. 
When all means are unknown, we find that the knowledge of the covariance matrix cannot be leveraged to better estimate the parameter collaboratively, even when sensors/agents know their variances as well as correlations.

Since we have multiple unknown parameters in this scenario, we formulate the distributed estimation problem as 
that of minimizing the CRB  (corresponding to the reciprocal of the Fisher information matrix). 
We then derive the optimal estimator whose variance matches the \rev{CRB.}

We show that, in \texttt{Scenario 2}, the optimal policy is to collect local observations and compute the sample mean.

\subsubsection{Bivariate (Two-sensor) Case}
As there are two unknown parameters, $\mu_1$ and $\mu_2$, the Fisher information takes a $2\times 2$ matrix form (called Fisher information matrix, FIM): 
\begin{align}
&\mathcal{I}(\mu_1, \mu_2)\notag\\
&\!= p_1 \mathcal{I}_{X_1}(\mu_1, \mu_2) + p_2 \mathcal{I}_{X_2} (\mu_1, \mu_2) +p_{1,2} \mathcal{I}_{(X_1, X_2)}(\mu_1, \mu_2)\notag\\
&\!= p_1\!\! \begin{bmatrix}
\frac{1}{\sigma_1^2} & 0\\
0&0
\end{bmatrix}
\!+ p_2\!\! 
\begin{bmatrix}
0&0\\
0&\frac{1}{\sigma_2^2}
\end{bmatrix}
\!+ p_{1,2}\!\!
\begin{bmatrix}
\frac{1}{(1-\rho_{1,2}^2)\sigma_1^2} & \frac{-\rho_{1,2}}{(1-\rho^2)\sigma_1\sigma_2}\\
\frac{-\rho_{1,2}}{(1-\rho_{1,2}^2)\sigma_1\sigma_2} & \frac{1}{(1-\rho_{1,2}^2)\sigma_2^2}
\end{bmatrix}\!.\notag 
\end{align}
The reciprocal of the FIM can be written as: 
\begin{align} \label{eq:inverse-two-mean}
&\mathcal{I}^{-1}(\mu_1, \mu_2) \notag\\
&= \frac{1}{\text{det}(\mathcal{I}(\mu_1, \mu_2))}
\begin{bmatrix}
\frac{p_2(1-\rho_{1,2}^2) + p_{1,2}}{(1-\rho_{1,2}^2)\sigma_2^2} & \frac{p_{1,2}\rho_{1,2}}{(1-\rho^2)\sigma_1\sigma_2} \\
\frac{p_{1,2}\rho_{1,2}}{(1-\rho^2)\sigma_1\sigma_2} & \frac{p_1(1-\rho_{1,2}^2) +p_{1,2}}{(1-\rho_{1,2}^2)\sigma_1^2}
\end{bmatrix},\\
&\text{det}(\mathcal{I}(\mu_1, \mu_2)) = \frac{p_{1,2}^2+p_1p_{1,2}+p_2p_{1,2}+p_1p_2(1-\rho_{1,2}^2)}{\sigma_1^2 \sigma_2^2 (1-\rho_{1,2}^2)}. \label{eq:deltaini}
\end{align}
The reciprocal of the FIM lower bounds the covariance matrix of any unbiased vector estimate of $\mu_1$, $\mu_2$. As sensor $S_1$'s objective is to estimate its corresponding mean parameter, $\mu_1$, with minimum variance, all it needs is to minimize the corresponding \rev{CRB},\footnote{If the objective of sensor $S_1$ was not only minimizing the variance of the estimate of $\mu_1$ but also helping sensor $S_2$ with the estimation of $\mu_2$, we may choose to optimize other statistics of the FIM, e.g., its determinant or the trace of its inverse, and consider not only transmission costs of receiving observations but also that of sending observations.  
We refer interested readers to the optimal design literature for different choices of optimization objectives. }
i.e., the $(1,1)$-entry of the reciprocal of the FIM, which lower bounds the variance of its estimate:
\begin{align}
    &\min\limits_{p_\emptyset, p_1, p_2, p_{1,2} \in [0,1]}\,\, \left(\mathcal{I}^{-1}(\mu_1, \mu_2)\right)^{1,1}\label{eq:unconstra21}\\
    &\,\,\text{s.t. }\,\, p_\emptyset + p_1 + p_2 +p_{1,2} = 1, \label{eq:unconstra22}\\
    &\quad\quad\, p_1 + (\alpha+1) p_{1,2} \leq E. \label{eq:two-mean-cons}
\end{align}
Different from the optimization problem in \texttt{Scenario 1}, the optimization objective (CRB) in~\eqref{eq:unconstra21} is not linear in $p_\emptyset, p_1, p_2, p_{1,2}$. 
While it is not as obvious as \texttt{Scenario 1}, the objective~\eqref{eq:unconstra21} is minimized when $p_2 = 0$; we defer this derivation to Appendix~\ref{sec:likelihood-scenario2-bivariate}.

\begin{figure}[t]
    \centering
    \includegraphics[width=0.5\linewidth]{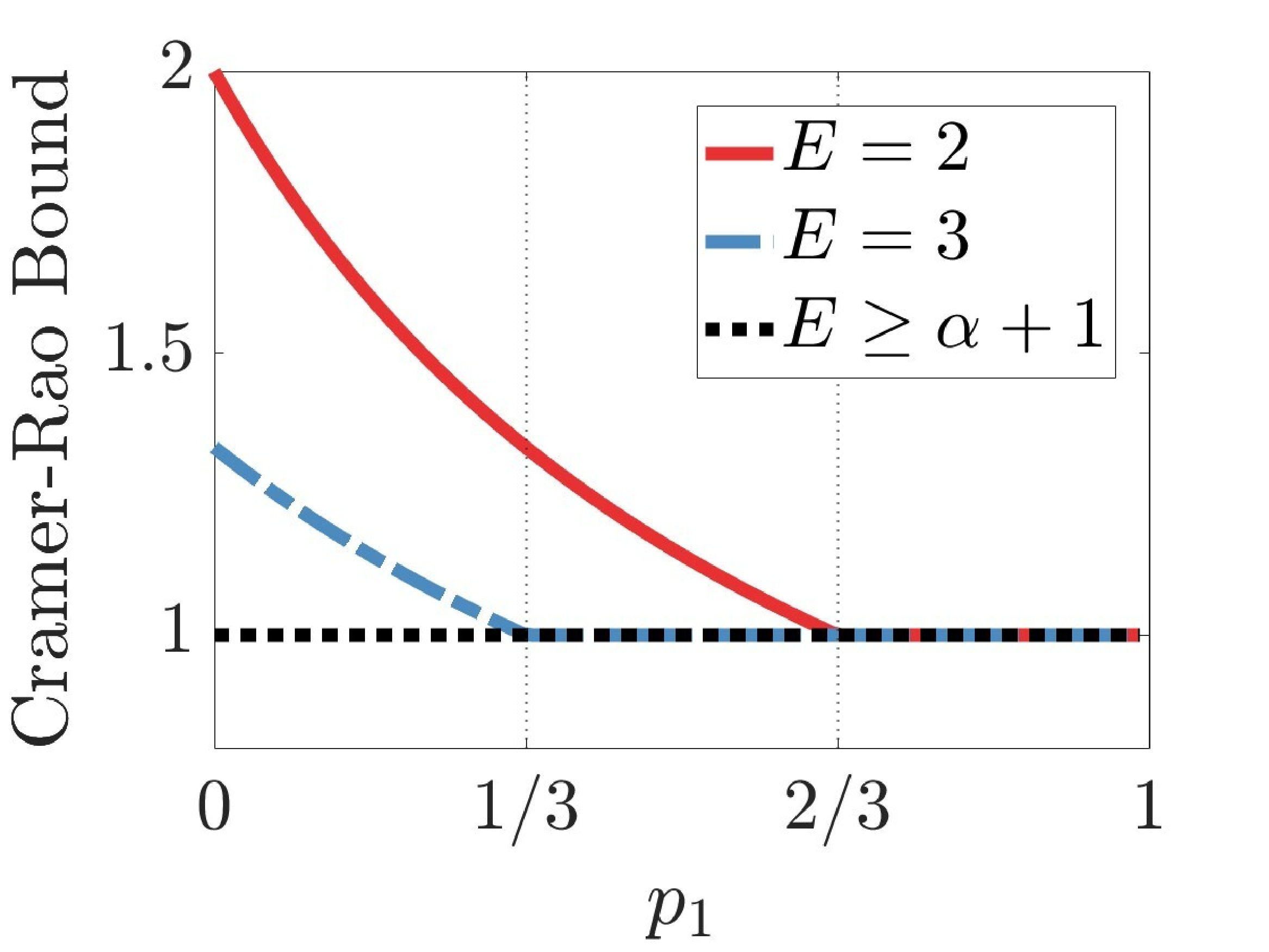} 
    \caption{\texttt{Scenario 2} Bivariate (Two-Sensor) Cases: (expected) Cramer-Rao bound under various static data collection policies (various values of $p_1$ and correspondingly $p_{1, 2} = \min\{1-p_{1}, (E-p_1)/(\alpha+1)\}$) when transmission cost $\alpha=3$, correlation $\rho_{1,2}=0.5$, and variance $\sigma_1^2 = \sigma_2^2 =1$}
    \label{fig:decentralized:two-mean-unknown-CRB}
\end{figure}

In Figure~\ref{fig:decentralized:two-mean-unknown-CRB}, we plot the values of the objective function~\eqref{eq:unconstra21} (i.e., the CRB) under various data collection strategies and resource constraints.
The black dotted line shows that in the absence of a resource constraint, i.e., resource constraint~\eqref{eq:two-mean-cons} is inactive, $E \geq \alpha + 1$, any value of $p_1 \in [0, 1)$ and correspondingly $p_{1,2} = 1- p_1$, $p_\emptyset = p_2 = 0$ yields the minimum CRB for $\sigma_1^2$. Surprisingly, the CRB cannot be decreased by using additional resources to collect bivariate observations; this implies that a bivariate joint observation $(x_1, x_2)$ is no more informative than a marginal observation $x_1$ in this scenario.  
When the resource constraint is active, $E<\alpha+1$, as illustrated by the blue and red lines, 
$p_1$ must be set large enough to achieve the same efficiency, i.e., one sample of $X_1$ every time slot, as unconstrained (black line), and minimize the CRB. 
Take the red line ($E=2$) as an example; when $p_1$ is large enough ($p_1 \geq 2/3$), $p_1 + p_{1,2} = 1$, the CRB is minimized; when $p_1 < 2/3$, $p_1 + p_{1,2} < 1$, allocating resources on sampling joint observations precludes making additional observations on $X_1$ and the CRB increases as $p_1$ decreases.

Assume we have
collected $N_1$ univariate samples of $X_1$, $N_2$ univariate samples of $X_2$ and $N_{1,2}$ bivariate samples of $(X_1, X_2)$. We then apply Kalman filtering to estimate $\mu_1$:
\begin{equation}\label{eq:kalman-two-mean-unknwon}
    \delta^* = g_1 \delta_1 + g_{1,2} \delta_{1,2}^\prime,
\end{equation}
where $\delta_1$ is the sample mean based on univariate observations, i.e., Eq.~\eqref{eq:delta_1}, and 
\begin{align}\label{eq:delta_12_prime}
    \delta_{1,2}^\prime = \frac{1}{N_{1,2}}\left(\sum_{i=1}^{N_{1,2}} x_{1,i} - \frac{N_2 \rho_{1,2}\sigma_1}{(N_2 + N_{1,2})\sigma_2}\left(x_{2,i} - \sum_{j=1}^{N_2} x_{2,j}\right)\right).
\end{align}
Note that the bivariate estimator~\eqref{eq:delta_12} does not apply here as $\mu_2$ is unknown. 
Interestingly, the  \rev{KFE given by}~\eqref{eq:kalman-two-mean-unknwon}, with optimal choice of $g_1, g_{1,2}$, is equivalent to the estimator given in~\cite{wilks1932moments} as well as the maximum likelihood estimator. We discuss this equivalence in more detail in Appendix~\ref{sec:likelihood-scenario2-bivariate}. 
It is also worth noting that $\delta_{1,2}^\prime$ and \rev{the KFE}, given by~\eqref{eq:kalman-two-mean-unknwon},   \rev{reduce} to the sample mean when $N_2 = 0$. Furthermore, as suggested by Figure~\ref{fig:decentralized:two-mean-unknown-CRB}, the optimal data collection policy and the corresponding estimator for this scenario is to use all resources to collect marginal observations of $X_1$ and use them to compute the sample mean.

\subsubsection{Multivariate (Multiple-sensor) Case}

With the insights obtained from the bivariate case above, we observe that, in \texttt{Scenario 2}, the Fisher information matrix associated with each type of sample can be computed a priori as in \texttt{Scenario 1}, but the Cramer-Rao bound minimization problem is more complicated than that in \texttt{Scenario 1}:
\begin{align}
    &\max\limits_{p_{\mathcal{K}}, \mathcal{K}\in \text{Pw}([K])} \,\, \left(\mathcal{I}^{-1}({\bm \mu})\right)^{1,1}, \\
&\,\,\text{s.t. }\,\,  \sum_{\mathcal{K}\in \text{Pw}([K])} p_{\mathcal{K}} = 1,\\
&\qquad\,\,\,p_1 +  \sum_{k\in[2, K]} (\alpha+1)p_{1,k}+ \sum_{k_1, k_2\in[2, K]} (2\alpha+1)p_{1, k_1, k_2}\notag\\
&\qquad\quad\,\,\,+...+((K-1)\alpha+1)p_{1,2,...,K} \leq E,
\end{align}
where 
\begin{align}
&\mathcal{I}({\bm \mu}) = p_1 \mathcal{I}_{X_1}({\bm \mu})+ \sum_{k\in[2, K]} p_{1,k}\mathcal{I}_{(X_1,X_k)}({\bm \mu})\notag\\ 
&\qquad\quad +\sum_{k_1, k_2\in[2, K]} p_{1, k_1, k_2}\mathcal{I}_{(X_1, X_{k_1}, X_{k_2})}({\bm \mu}) +...\notag\\
&\qquad\quad +p_{1,2,...,K}\mathcal{I}_{(X_1, X_2,..., X_K)}({\bm \mu}).
\end{align}
Despite the more complicated optimization problem, we find that allocating resources to generating joint samples or samples of other variables does not provide more information than is obtained by allocating the same amount of resources to the collection of local marginal samples. We show in Appendix~\ref{sec:likelihood-scenario2-multivariate} that, for the multivariate case, the sample mean is also the maximum likelihood estimator for this scenario. 
It is worth noting that setting the $(1,1)$-entry of FIM as our optimization objective and aiming to find the optimal unbiased estimator plays a key role in our finding that leveraging collaboration to produce an estimate with a smaller variance is not possible. We leave the extension of our analysis to other optimization objectives or biased estimators, accounting for two-stage estimators~\cite{brogan1978estimating}, preliminary tests~\cite{saleh2006theory} or James-Stein estimators~\cite{casella2002statistical}, for future work.


\section{\myhl{\texttt{Scenario 3} - Correlation Information Unavailable}}\label{sec:correlation-unknown}

In this section, we relax the   assumption that the correlation parameters are known a priori; for example, if sensors/agents have different modalities, it may be difficult to know the correlations beforehand. 
We still consider the distributed problem from the perspective of a newly deployed sensor/agent and assume that old sensors/agents have learned their corresponding parameters fairly well.

In contrast to \texttt{Scenario 1} and \texttt{Scenario 2}, the Fisher information contained within various types of joint samples is initially unknown since the correlation parameters are unknown. \myhl{Through joint observations collected over time,} we can estimate these correlation parameters and thereby estimate the Fisher information for each type of joint sample. Our correlation and Fisher information estimates improve as we increase the number of \myhl{joint observations}. However, ultimately, our objective is to maximize the Fisher information we obtain regarding $\mu_1$ (minimizing the variance of our estimate of $\mu_1$) instead of obtaining increasingly accurate estimates of the correlation parameters or Fisher information.

\rev{Hence, we face an exploration-exploitation dilemma each time we select the type of sample to collect. Should we gather a joint observation to improve our Fisher information estimates (exploration), or should we rely on our current estimates and choose the sample that maximizes such   Fisher information estimates (exploitation)?}

\subsection{Estimation as a Multi-Armed Bandit Problem}

The exploration-exploitation dilemma in our distributed parameter estimation problem in \texttt{Scenario 3} bears some resemblance to the classical stochastic  MAB  problem. In the classical stochastic MAB problem, there is a set of arms, and each arm is associated with a distribution; at each decision round, the learner chooses one arm and obtains feedback sampled from the chosen arm's distribution. In this scenario, a set of sensor(s) can be modeled as one arm, and our data collection policy chooses one arm (set of sensor(s)) to collect a sample from at each decision round.

In Section~\ref{sec:MAB-model}, we introduce the MAB model for our distributed estimation problem, where we make two adjustments to our problem formulation to adapt it to the MAB model. First, we confine ourselves to collecting either univariate marginal observations or bivariate joint observations, even in the multivariate (multiple-sensor) setting. 
This modification not only prevents the exploration space from becoming combinatorially large but also allows us to use the critical thresholds (e.g.,~\eqref{eq:condikey}) derived in Section~\ref{sec:scenario-1-bivariate}, rather than nonlinear thresholds as illustrated in Figure~\ref{fig:trivariate}, to prioritize the collection of different types of samples.
This simple and intuitive prioritization scheme makes it straightforward to design adaptive data collection policies. Second, we map time slots in our sensor/agent system to decision rounds in the MAB model. Depending on the resource budget and transmission cost, the MAB model of our distributed estimation problem enforces different granularities of decision rounds to maintain compliance with the resource constraint.

On the other hand, our distributed parameter estimation problem in this scenario also differs fundamentally from a classical stochastic MAB problem. This fundamental difference lies in the feedback and the optimization objectives. Specifically, the feedback is itself the reward objective that an MAB policy aims to maximize in a classical stochastic MAB problem, while the feedback in our parameter estimation problem is not our maximization objective. In fact, our maximization objective, the Fisher information about $\mu_1$, can only be inferred from the feedback samples. Hence, existing stochastic MAB policies cannot be directly applied to our problem without carefully designing surrogate rewards. 
In the following, we propose two kinds of surrogate rewards tailored specifically to our parameter estimation task and incorporate them into popular MAB policies to form four sequentially adaptive data collection policies.

\subsection{Reward Maximization and Estimator} \label{sec:scenario3-opt-problem}

In the following, we first formulate our Fisher information maximization objective in \texttt{Scenario 3} as a cumulative reward maximization multi-armed bandit model. Then, we present an estimator that can utilize samples collected by adaptive data collection policies.

The Fisher information maximization problem in \texttt{Scenario 3} is similar to that in \texttt{Scenario 1} (except correlations are unknown, and we restrict ourselves to only collect univariate or bivariate samples). This is because the parameter of interest, $\mu_1$, is orthogonal to the other unknown parameters, $\rho_{1, k},$ for $ k\in[2, K]$, which makes the Fisher information matrices (FIM) diagonal, 
\begin{align}
   \mathcal{I}_{X_1}(\mu_1) &= \begin{bmatrix}
       \frac{1}{\sigma_1^2} & 0\\
        0 & 0
   \end{bmatrix},\\
   \mathcal{I}_{(X_1, X_k)}(\mu_1, \rho_{1,k}) &= \begin{bmatrix}
       \frac{1}{(1-\rho_{1,k}^2)\sigma_1^2} & 0\\
        0 & \frac{1+\rho_{1,2}^2}{(1-\rho_{1,k}^2)^2}
   \end{bmatrix}.
\end{align}
Hence, the objective of sensor $S_1$, to minimize the $(1,1)$-entry of the reciprocal of the cumulated Fisher information matrix (FIM), modeled as a cumulative reward maximization problem, is given by
\begin{align}
    &\max\limits_{\substack{p_1{(\tau)}, p_{1, j}{(\tau)} \in \{0,1\},\\ \forall j\in[2, K], \forall {\tau} \in [\mathcal{T}]}} \sum_{\tau = 1}^{\mathcal{T}} p_1{(\tau)}\frac{c}{\sigma_1^2} + \sum_{j=2}^K p_{1,j}{(\tau)}\frac{1}{(1-\rho_{1, j}^2)\sigma_1^2},\label{eq:scenario-3-begin}
    \end{align}
    \begin{align}
    &\,\,\text{s.t. }\,\,p_1{(\tau)} + \sum_{j=2}^K p_{1,j}{(\tau)} = 1, \forall \tau \in [\mathcal{T}],\label{eq:scenario-3-mid}\\
    &\quad\quad\, c = \begin{cases}
        1, & \text{ if } E > \alpha + 1,\\
        \lceil(\alpha+1)/E\rceil, & \text{ if } 1 \leq E < \alpha + 1,\\
        \lfloor(\alpha + 1)\rfloor, & \text{ if } E < 1,
    \end{cases}\label{eq:scenario-3-end}
\end{align}
where $c/\sigma_1^2$ is the reward corresponding to arm $1$ and $1/((1-\rho_{1, j}^2)\sigma_1^2)$ is the reward corresponding to arm $j$. Note that the optimization problem described by~\eqref{eq:scenario-3-begin}-\eqref{eq:scenario-3-end} cannot be solved offline as correlations $\rho_{1, j}, j\in [2, K]$ are unknown and therefore the objective function is initially unknown in this scenario.

To estimate the parameter of interest, $\mu_1$, with $N_1$ univariate samples of $X_1$ and $N_{1,j}$ bivariate samples of $(X_1, X_j)$, $j \in [2, K]$, we apply a  Kalman filter:
\begin{equation}\label{eq:kalman-rho-unknown}
    \delta^* = g_1 \delta_1 + \sum_{j=2}^K g_{1,j} \delta_{1,j}^{\prime\prime},
\end{equation}
where the univariate estimator $\delta_1$ is the sample mean of univariate observations,~\eqref{eq:delta_1}, and the bivariate estimator is as proposed in~\cite{bishwal2008note},
\begin{align}
    &\delta_{1,j}^{\prime\prime} = \bar{x}_1 - \hat{\beta}_j(\bar{x}_j-\mu_j), & \hat{\beta}_j = \frac{\sum_{i}(x_{1,i}-\bar{x}_1)(x_{j,i}-\bar{x}_j)}{\sum_{i}(x_{j,i} - \bar{x}_j)^2}.
\end{align}
Estimator $\delta_{1,j}^{\prime\prime}$ does not achieve the Cramer-Rao bound but is unbiased and more efficient than the sample mean when $|\rho_{1,j}|  > 1/\sqrt{N_{1,j}-2}$. We refer interested readers to~\cite{bishwal2008note} for details. Note that bivariate estimators~\eqref{eq:delta_12} and~\eqref{eq:delta_12_prime} are not applicable here as correlations $\rho_{1, j}, j\in [2, K]$ are unknown. For the same reason, the optimal weights for the \rev{KFE} cannot be computed a priori. Based on the intuition that the estimator in which we have more confidence should be given greater weight, we apply the following heuristic: $g_1 = N_1/N$, $g_{1, j} = N_{1, j}/N$, $\forall j \in [K] \setminus \{1\}$, where $N = N_1 + \sum_{j=2}^K N_{1,j}$.

\subsection{Surrogate Reward Design}\label{sec:surrogate}

As discussed in previous subsections, classical multi-armed bandit algorithms cannot be directly applied to \texttt{Scenario~3} to solve our data collection and collaboration problem since the rewards (true Fisher information) corresponding to sampling bivariate observations are unknown.
In the following, we introduce two kinds of surrogate rewards, the Fisher information estimate and $z$-transformed correlation estimate, selected particularly for our distributed parameter estimation problem. 
\myhl{Note that both surrogate rewards can be computed in constant time during 
each decision round~\cite{chan1983algorithms,ling1974comparison}.}

\subsubsection{Fisher Information Estimate}
As our maximization objective is the true Fisher information, a straightforward choice for surrogate reward is the observed/estimated Fisher information, i.e., 
\begin{align}
&\hat{\mathcal{I}}_{(X_1, X_j)}(\mu_1) = \left((\hat{{\bm \Sigma}}_{(X_1, X_j)})^{-1}\right)^{1,1} = \frac{1}{(1-(\hat{\rho}_{1,j})^2)\sigma_1^2}, \label{eq:estimated-rho1}
\end{align}
\begin{align}
&\hat{\rho}_{1,j} = \frac{\sum_{i}(x_{1,i} - \bar{x}_1)(x_{j,i} - \bar{x}_j)}{\sqrt{\sum_{i}(x_{1,i} - \bar{x}_1)^2}\sqrt{\sum_{i}(x_{j,i} - \bar{x}_j)^2}}.\label{eq:estimated-rho}
\end{align}

\subsubsection{$z$-transformed Correlation Estimate}

Notice that a bivariate sample $(x_1, x_j)$ contains more Fisher information than a univariate sample $x_1$ if $\rho_{1,j} > \sqrt{\alpha/(1+\alpha)}$ and contains more Fisher information than another bivariate sample \myhl{$(x_1, x_k)$} when $\rho_{1,j} > \rho_{1, k}$ (see, e.g., Figure~\ref{fig:trivariate}). \myhl{Motivated by this observation,}
another option for surrogate reward is to utilize the estimated correlations. Specifically, we use the Fisher's $z$-transformation~\cite{fisher1921014} of the sample correlation coefficient as the surrogate reward of a bivariate sample $(x_1, x_j)$, 
\begin{equation}\label{eq:z-trans-rho}
   z_{1,j} = \text{tanh}^{-1}(\hat{\rho}_{1,j}) = \frac{1}{2}\frac{1+\hat{\rho}_{1,j}}{1-\hat{\rho}_{1,j}},
\end{equation}
where $\hat{\rho}_{1,j}$ is computed using~\eqref{eq:estimated-rho}; and we use $z_1 = \text{tanh}^{-1}(\sqrt{\alpha/(1+\alpha)})$ as the surrogate reward for univariate sample $x_1$.
We propose to use the $z$-transformation correlation estimate instead of simply the sample correlation because the distribution of sample correlation is highly skewed (which many classical bandit algorithms do not handle) when the true value of correlation, a.k.a., population correlation, is close to one~\cite{fisher1915frequency}.
On the other hand, Fisher's $z$-transformation of sample correlation is approximately normally distributed and has a variance that is stable over different values of population correlation~\cite{fisher1946statistical, cramer1999mathematical}.

\begin{figure*}[t]
   \centering
        \subfloat[all correlations are very small]{\includegraphics[width=0.24\linewidth]{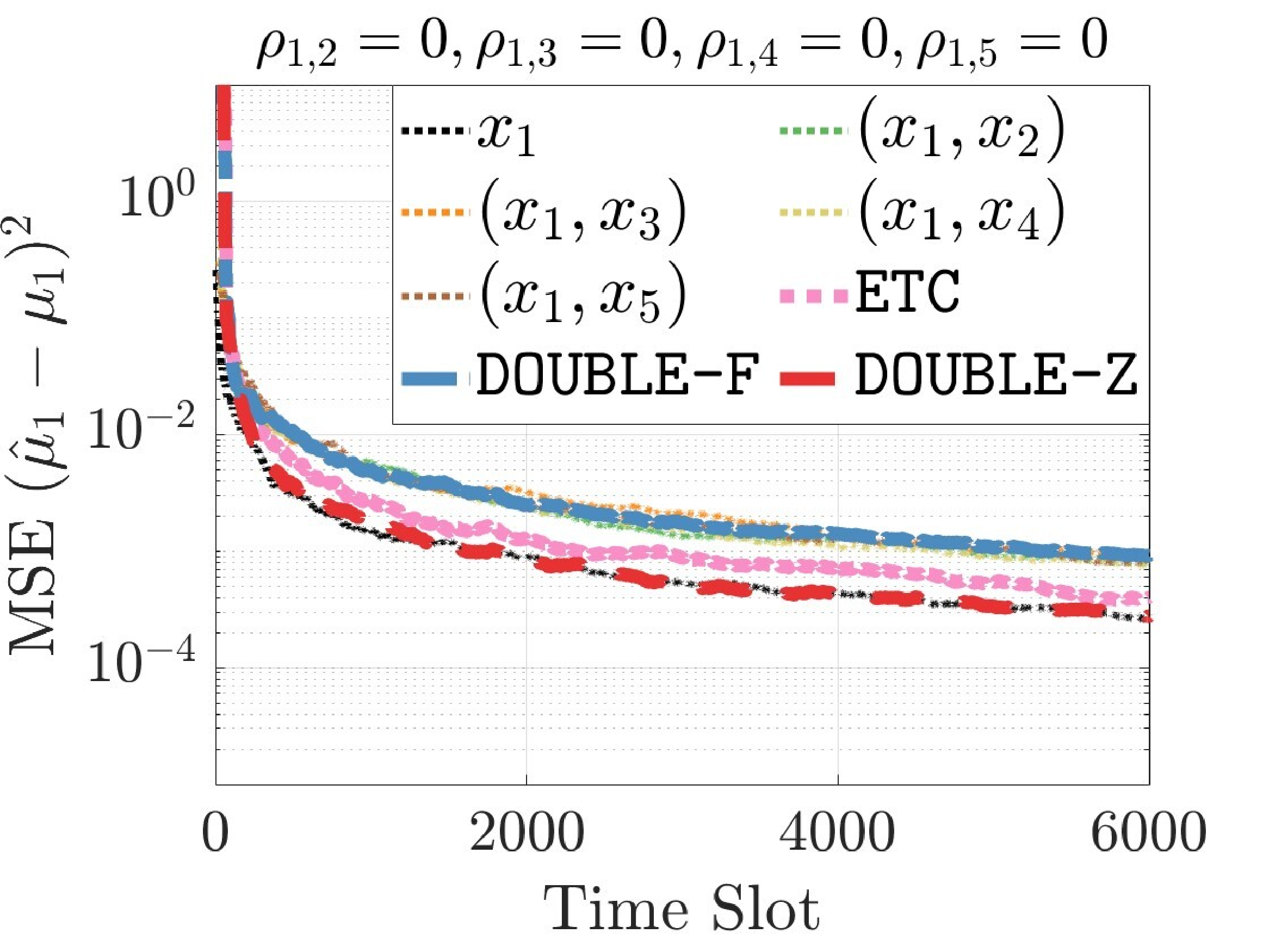}\label{subfig:a}}
        \hfil
        \subfloat[all correlations are small]{\includegraphics[width=0.24\linewidth]{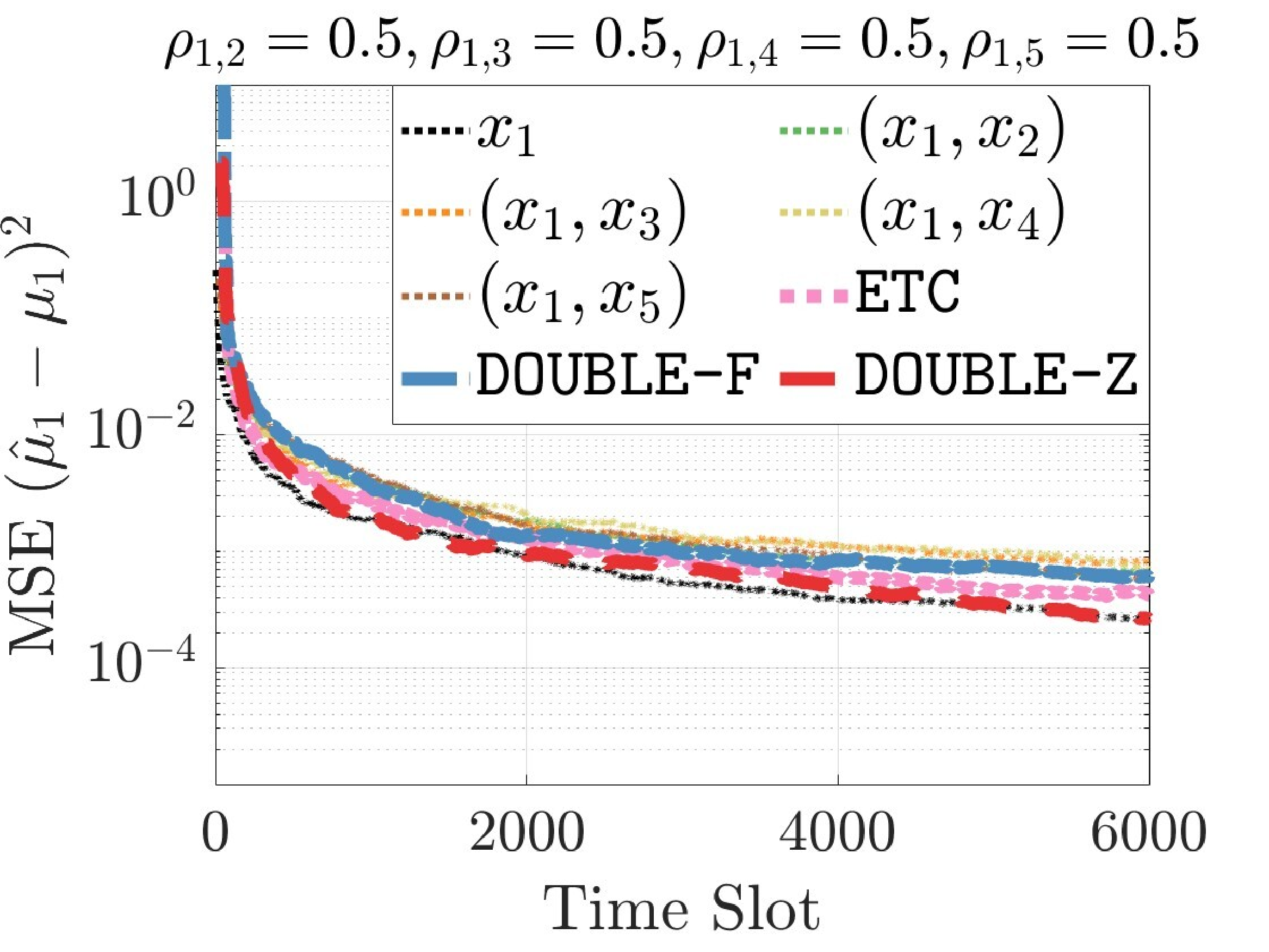}\label{subfig:b}}
        \hfil
        \subfloat[one correlation is large]{\includegraphics[width=0.24\linewidth]{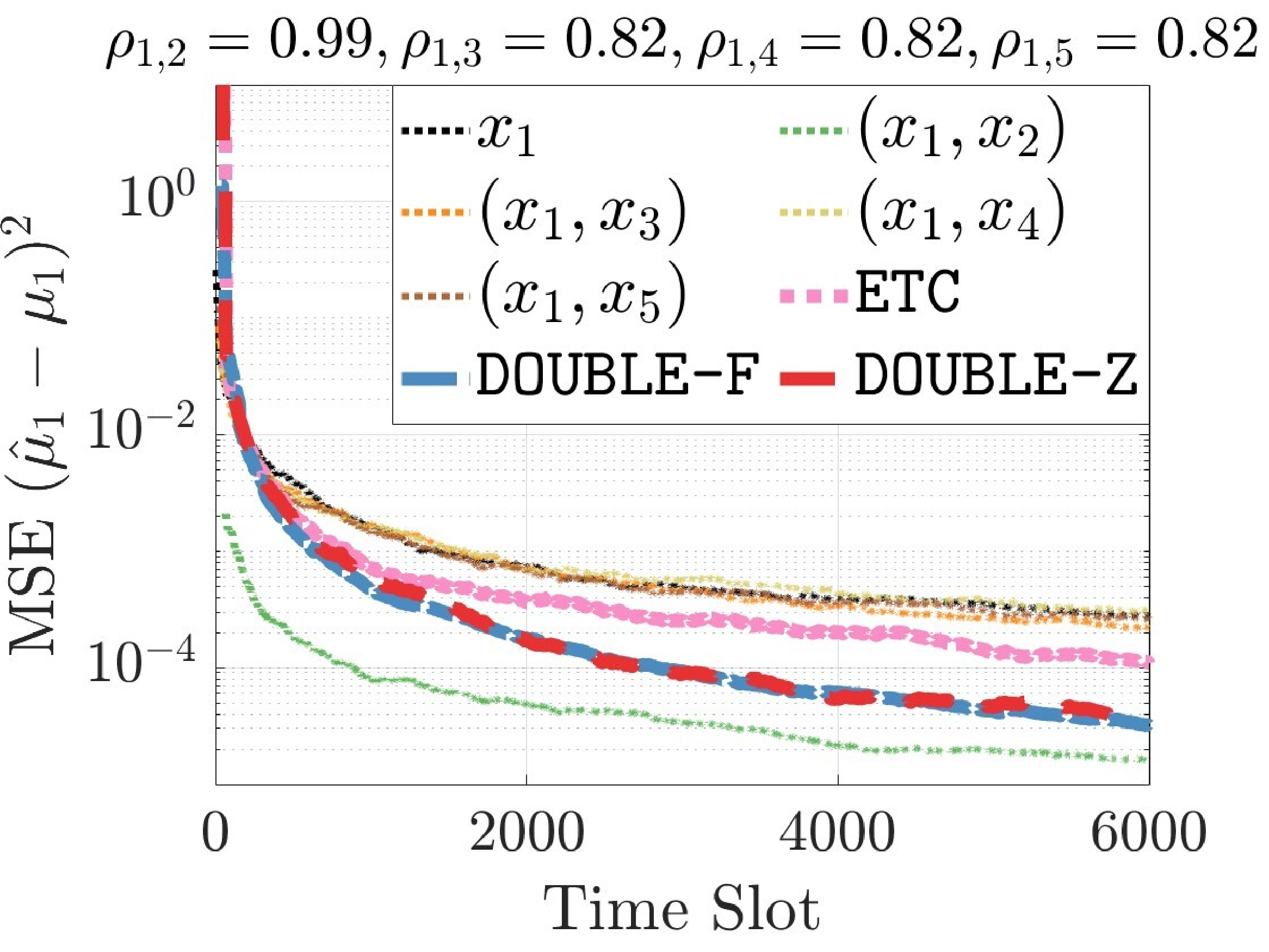}\label{subfig:c}}
        \hfil
        \subfloat[all correlations are large]{\includegraphics[width=0.24\linewidth]{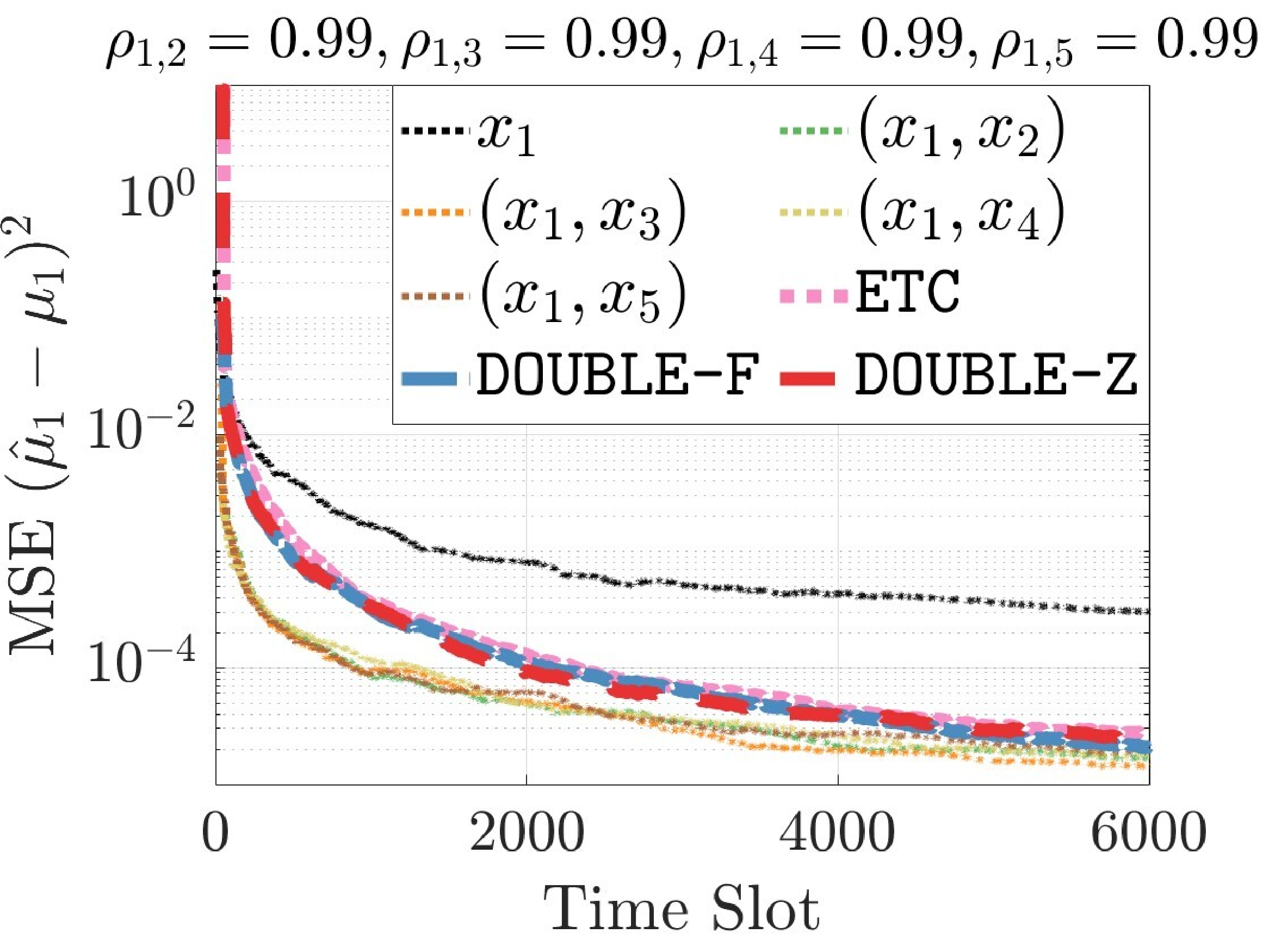}\label{subfig:d}}
        \hfil
        
        \subfloat[all correlations are very small]{\includegraphics[width=0.24\linewidth]{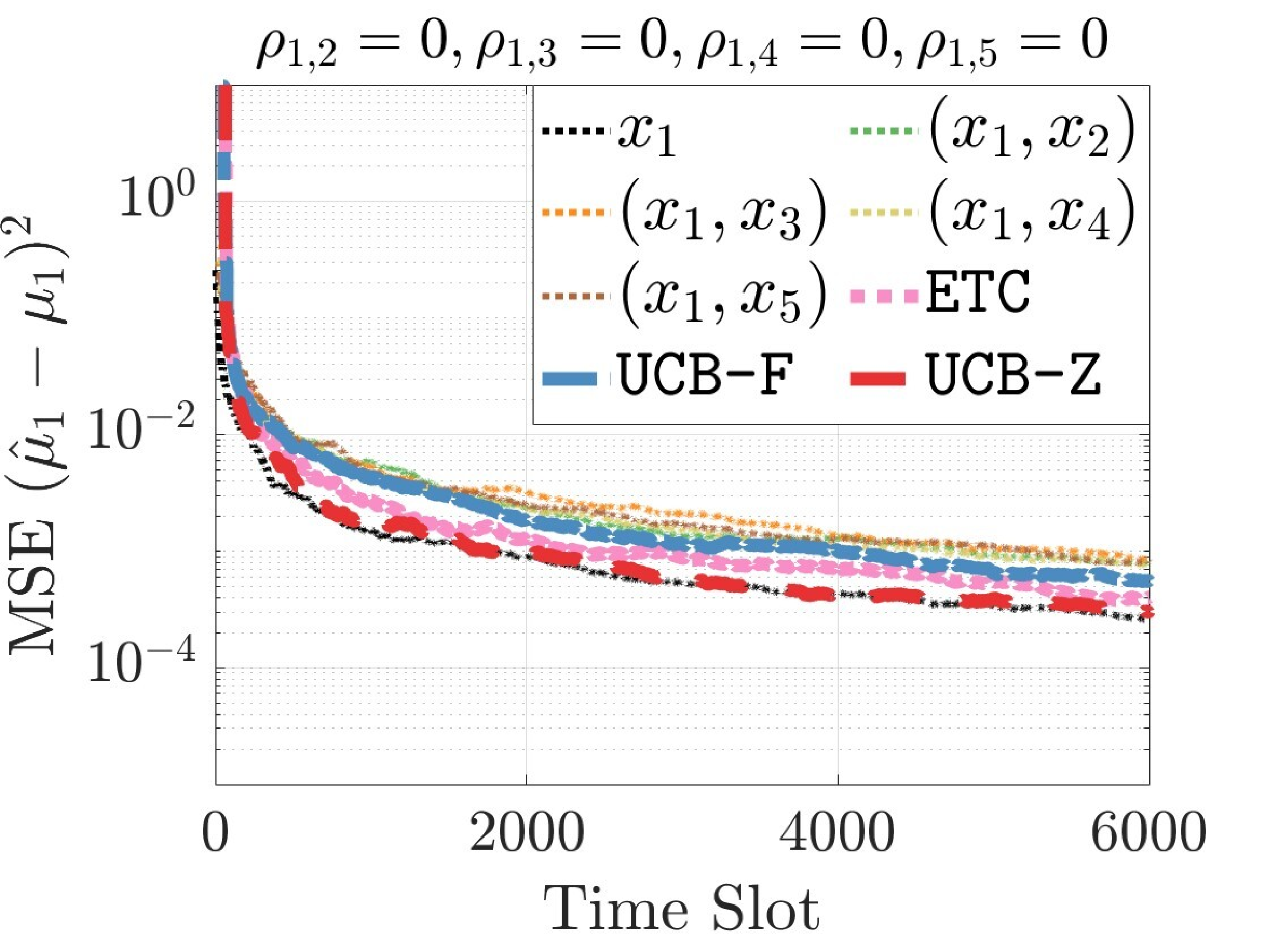}\label{subfig:e}}
        \hfil
        \subfloat[all correlations are small]{\includegraphics[width=0.24\linewidth]{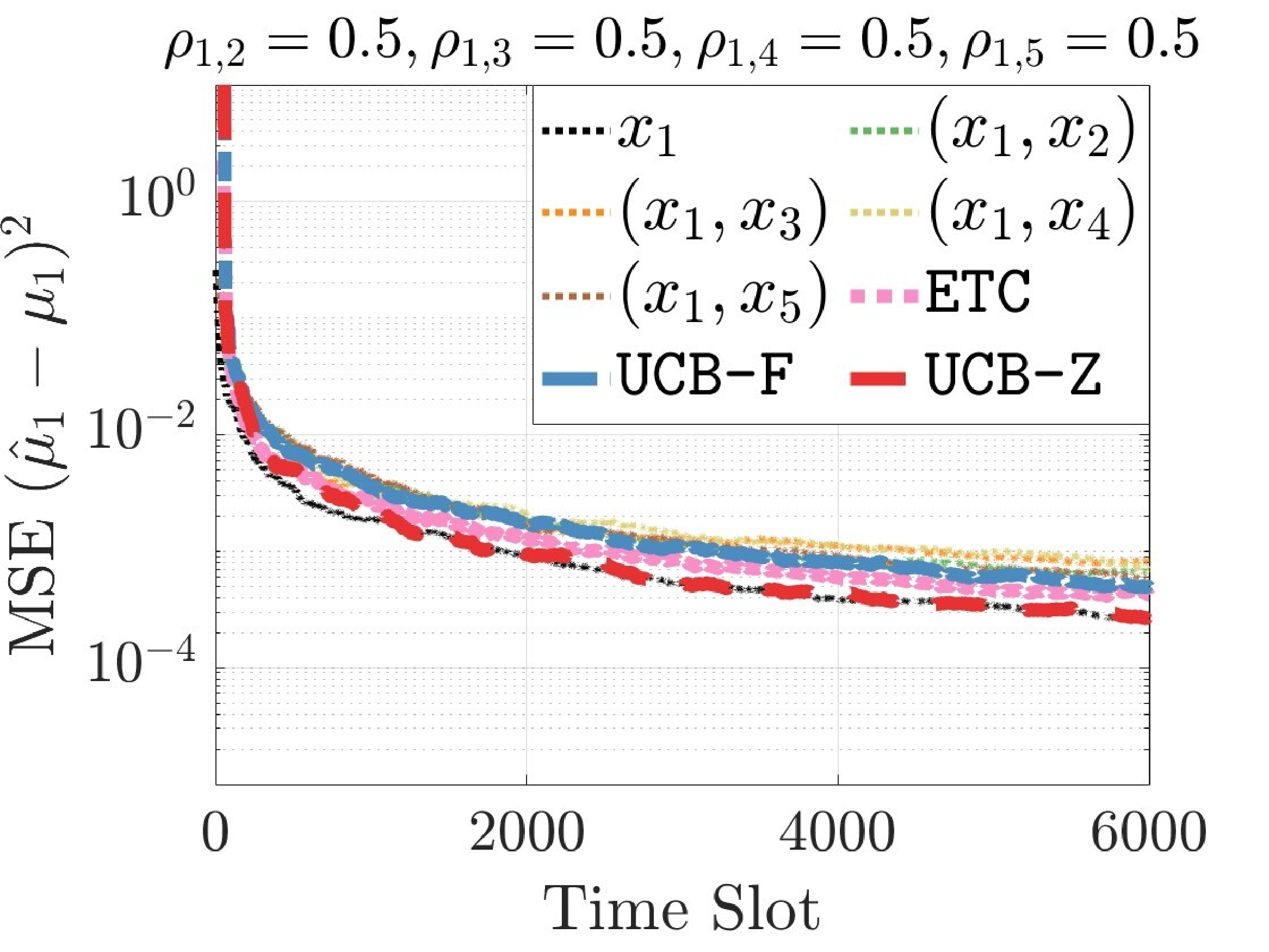}\label{subfig:f}}
        \hfil
        \subfloat[one correlation is large]{\includegraphics[width=0.24\linewidth]{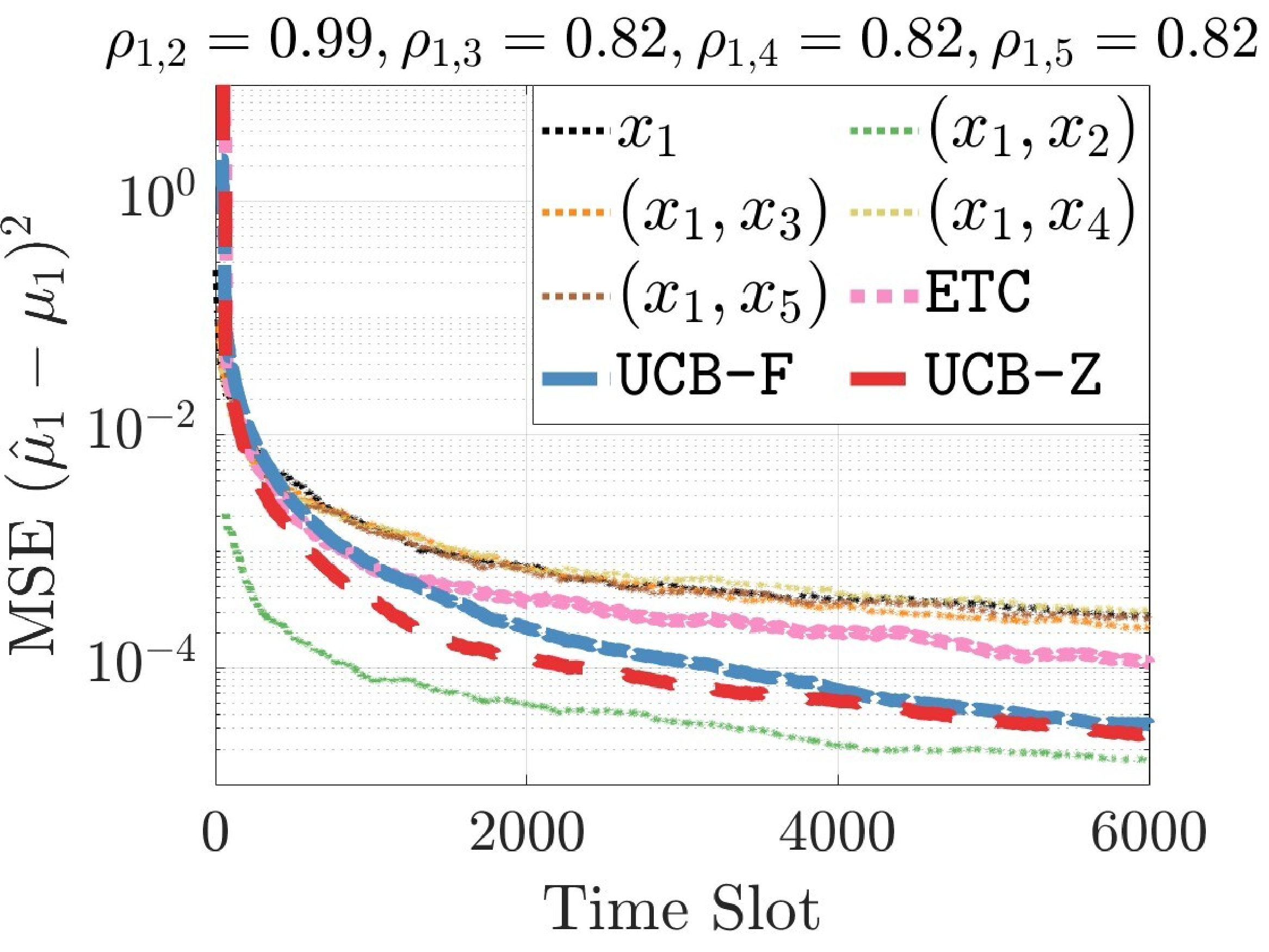}\label{subfig:g}}
        \hfil
        \subfloat[all correlations are large]{\includegraphics[width=0.24\linewidth]{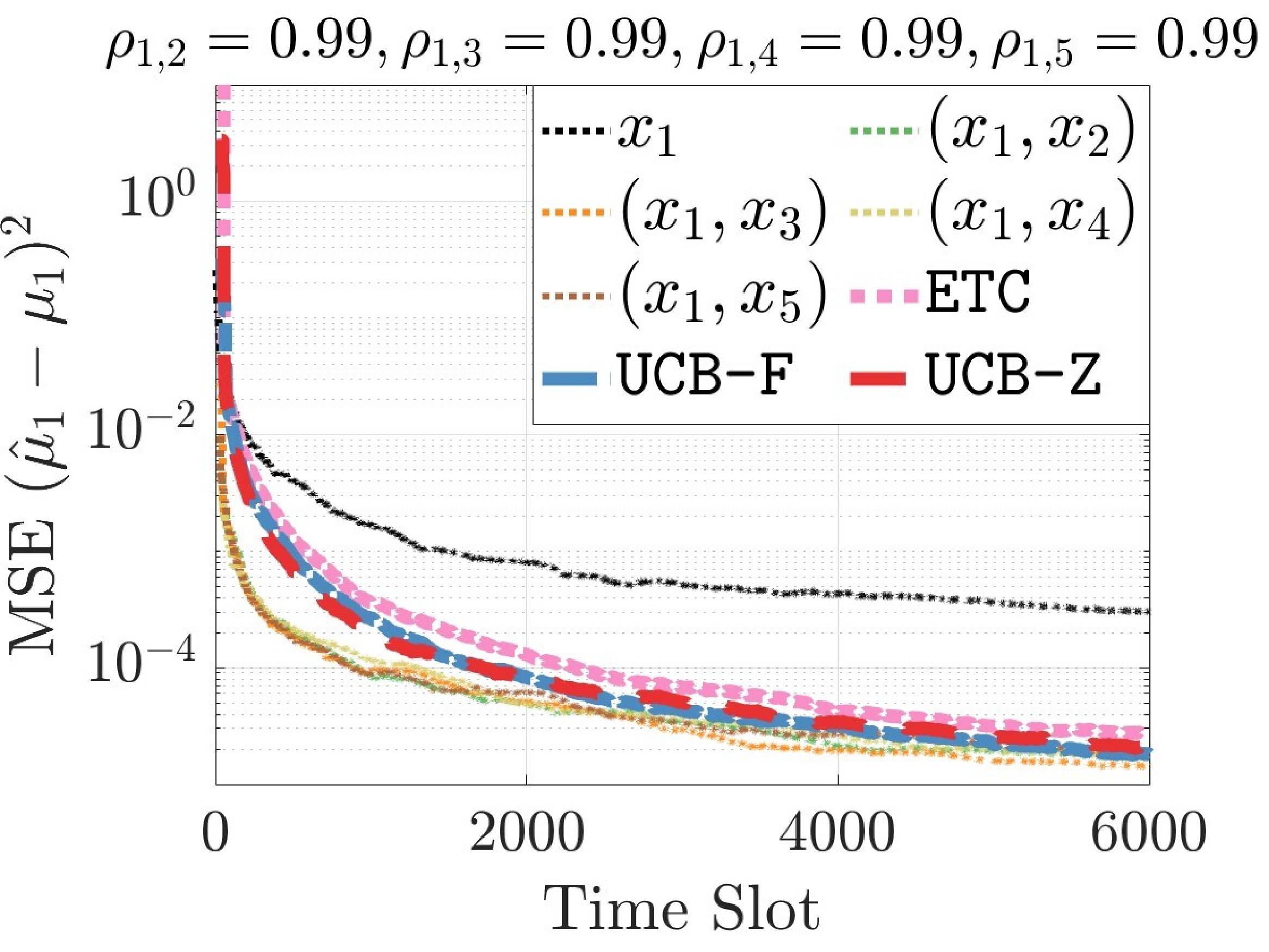}\label{subfig:h}}
        
        \caption{\texttt{Scenario 3} Multivariate (Multiple-Sensor) Cases: mean squared error of estimate over time when $\alpha = 2$, $E=0.6$, means and variances are set to $1$, and (a)(e) all correlations are very small (far below threshold $\approx 0.82$); (b)(f) correlations are increased, getting closer but still below the threshold; (c)(g) correlations increased with respect to previous setup, one of the correlations is far above the threshold; (d)(h) all correlations are far above the threshold. }\label{fig:scenario3-5-variate}
\end{figure*}

\subsection{Incorporating MAB policies with Surrogate Reward}

In the following, we incorporate the above surrogate rewards into two types of MAB policies to form four sequential adaptive data collection policies for our parameter estimation problem. Pseudo codes for the proposed policies are given in Appendix~\ref{sec:pseudocode}. 
\myhl{Note that each proposed adaptive policy only requires a constant number of computations during each decision round. Hence, their computational complexities are linear in the length of the time horizon.}

\subsubsection{Doubling-Trick (\texttt{DOUBLE}) Based Policies}\label{sec:doubling-based}
The doubling-trick-based policies alternate between exploration and exploitation repeatedly, with the exploitation period continually increasing in length.
This kind of control policy was first proposed by Chernoff~\cite{chernoff1959sequential} for the sequential design of experiments.
We incorporate a doubling-trick-based policy with the Fisher information estimate and the $z$-transformed correlation estimate surrogate rewards, leading to \texttt{DOUBLE-F} and \texttt{DOUBLE-Z}, respectively.
Specifically, we let the set of time slots dedicated to exploration to lie in
\begin{align}
    &\mathcal{H}_\eta \equiv [2K] \cup \{\lceil \eta^\ell\rceil: \ell\in\mathbb{N}\}, & \eta > 1.
\end{align}
When ${\tau} \in \mathcal{H}_\eta$, we explore by pulling an arm uniformly at random. When ${\tau} \notin \mathcal{H}_\eta$, we exploit by pulling the arm with the largest surrogate reward, i.e.,
\begin{align}
    J_{\tau} = \arg\max_{j\in [K]}\begin{cases}
        \hat{\mathcal{I}}_{(X_1, X_j)}(\mu_1), & \text{ under } \texttt{DOUBLE-F},\\
        z_{1,j}, & \text{ under } \texttt{DOUBLE-Z}.
    \end{cases}
\end{align}

\subsubsection{Upper Confidence Bound (\texttt{UCB}) Based Policies}\label{sec:UCB-based}

The UCB-based policies balance exploration and exploitation by maintaining a confidence interval for each arm centered on its empirical mean and always selecting the arm with the highest upper confidence bound~\cite{lai1985asymptotically}. 
We form policies \texttt{UCB-F} and \texttt{UCB-Z} by incorporating a UCB-based policy with the Fisher information estimate and the $z$-transformed correlation estimate surrogate rewards, respectively. 
Specifically, at each decision round, we pull arm $J_{\tau}$ such that
\begin{align}
    J_{\tau} = \arg\max_{j\in [K]}\begin{cases}
    \hat{\mathcal{I}}_{(X_1, X_j)}(\mu_1) + \texttt{CI}(j),& \text{ under } \texttt{UCB-F},\\
    z_{1,j} + \texttt{CI}(j),& \text{ under } \texttt{UCB-Z},
    \end{cases}
\end{align}
where the confidence interval width $\texttt{CI}(j)$ is defined, with policy parameter $a$ and $\epsilon$, as 
\begin{equation}\label{eq:confidence-interval}
    \texttt{CI}(j) \equiv \sqrt{\frac{a \log \epsilon^{-1}}{2 n_{1, j}}}.
\end{equation}

\subsection{Numerical Study}~\label{sec:numerical}

We experimentally evaluate the proposed adaptive policies, \texttt{DOUBLE-F}, \texttt{DOUBLE-Z}, \texttt{UCB-F}, and \texttt{UCB-Z}, on $5$-variate cases in \texttt{Scenario 3}, where the unknown correlations are varied from all being very small (wherein independent learning is more efficient) to all being very large (wherein collaboration is more efficient). 
\myhl{
We compare our policies with five static policies and a baseline adaptive policy. Specifically, the five static policies always sample $ x_1$, $ (x_1, x_2) $, $ (x_1, x_3) $, $ (x_1, x_4) $, and $ (x_1, x_5) $, respectively. The baseline adaptive policy leverages the exploration-then-commit (\texttt{ETC}) idea~\cite{perchet2016batched}, sampling joint observations from the beginning (exploration) to a pre-determined point, time slot $100$, and directly applying the sequential correlation significance testing to decide whether to continue sampling joint observations or to switch to only sample univariate observations (exploitation).   
}
We report the mean squared errors of the estimates (using~\eqref{eq:kalman-rho-unknown}) over time for all policies.
The reported values are averaged over $100$ independent \rev{simulation runs.}

Figure~\ref{fig:scenario3-5-variate} presents the experimental results. Figure~\ref{fig:scenario3-5-variate} shows that \texttt{DOUBLE-Z} and \texttt{UCB-Z} successfully learn the optimal data collection and collaboration strategies in all four correlation settings. Although \texttt{DOUBLE-F} and \texttt{UCB-F} perform slightly better than \texttt{DOUBLE-Z} and \texttt{UCB-Z} when all correlations are large and collaboration should be prioritized, they fail to learn that sampling local observations is better when correlations are small (as shown in Figures~\ref{subfig:a},~\ref{subfig:e},~\ref{subfig:b}, and~\ref{subfig:f}).

\myhl{
The performance of \texttt{ETC} relies on setting an appropriate length of exploration period. However, the length depends on how close the correlations are to each other and to the threshold and the length of the overall learning horizon, which are often unknown beforehand in real applications. Indeed, Figures~\ref{subfig:c} and~\ref{subfig:g} show that \texttt{ETC} performs poorly as the exploration length is too short for this problem instance.   The four proposed policies behave similarly, eventually approaching the optimal policy, which consists of taking joint samples of the highly correlated variables $(x_1, x_2)$.
}

\section{Conclusion And Future Direction}\label{sec:conclusion}
\myhl{ This work considers a resource-constrained sequential distributed parameter estimation problem with vertically partitioned data, where an agent's objective is to estimate a local unknown parameter. We seek optimal resource allocations between local data collection and collaborative data transfer under three scenarios.}
In \texttt{Scenario 1}, we identify the conditions on correlations and data transmission cost under which collaboration is more efficient than independent estimation. In \texttt{Scenario 2}, we found that collaboration never leads to more efficient estimation. 
\myhl{
In \texttt{Scenario 3}, optimal resource allocation cannot be computed offline due to the lack of knowledge about the correlations. Hence, we model the problem as a multi-armed bandit and propose novel surrogate reward designs to adopt multi-armed bandit policies. 
}

This work opens up a number of directions for future research.
\myhl{
One direction is to extend the problem to the case where data transmission is more expensive than reception. The extra constraints due to the cost difference may slightly change the conditions under which collaboration is more efficient than independent estimation.} 
Another direction is to extend our multi-armed bandit model for \texttt{Scenario 3} to allow sampling multivariate observations in multivariate cases. This extension would require more involved surrogate rewards design or the adoption of ideas of combinatorial/multi-play multi-armed bandit. 
Other future directions include considering the resource allocation problem for classification rather than estimation purposes under a parametric or non-parametric model.

\appendices


\section{Maximum Likelihood Estimators (MLEs)}\label{sec:likelihood}

\subsection{The MLE in \texttt{Scenario 1} Bivariate Case}\label{sec:likelihood-scenario1}

In the following, we derive the MLE of the mean parameter $\mu_1$ in the bivariate Gaussian distribution, under \texttt{Scenario 1}. 
First, we introduce some notation:
\begin{itemize}
\item 
 sample means   $\bar{x}_1$ and $\bar{x}_2$, sample variances    $\hat{\sigma}_1$ and $\hat{\sigma}_2$,    sample correlation   $\hat{\rho}_{1,2}$ are computed from $n_{1,2}$ bivariate samples of $(X_1, X_2)$
 \item  sample mean $\bar{x}_1^\prime$ and sample variance $\hat{\sigma}_1^\prime$  are computed from $n_1$ univariate samples of $X_1$
 \item sample mean $\bar{x}_2^\prime$ and sample variance $\hat{\sigma}_2^\prime$ are computed from $n_2$ univariate samples of $X_2$. 
 \end{itemize} 
As noted by  Wilks~\cite{wilks1932moments}, given a bivariate Gaussian distribution with parameters $\mu_1, \mu_2, \sigma_1, \sigma_2, \rho_{1,2}$, the joint distribution of the above estimators can be written from several well-known independent distributions as
\begin{align}
    &\mathcal{F}(\bar{x}_1, \bar{x}_2, \hat{\sigma}_1, \hat{\sigma}_2, \hat{\rho}_{1,2}, \bar{x}_1^\prime, \hat{\sigma}_1^\prime, \bar{x}_2^\prime, \hat{\sigma}_2^\prime |\mu_1, \mu_2, \sigma_1, \sigma_2, \rho_{1,2})\notag\\
    &= \mathcal{N}\left(\bar{x}_1, \bar{x}_2|\mu_1, \mu_2, \frac{{\bm\Sigma}}{n_{1,2}}\right)\cdot \\
    &\qquad\mathcal{W}\left(\hat{\sigma}_1, \hat{\sigma}_2, \hat{\rho}_{1,2}| n_{1,2}-1, \frac{{\bm\Sigma}}{n_{1,2}}\right) \cdot \notag\\
    &\qquad\mathcal{N}\left(\bar{x}_1^\prime|\mu_1, \frac{\sigma_1}{n_1}\right) \cdot \mathcal{X}^2\left( \hat{\sigma}_1^\prime| \frac{n_1-1}{2}, \frac{2\sigma_1^2}{n_1}\right) \cdot\notag\\
    &\qquad\mathcal{N}\left(\bar{x}_2^\prime|\mu_2, \frac{\sigma_2}{n_2}\right) \cdot \mathcal{X}^2\left(\hat{\sigma}_2^\prime| \frac{n_2-1}{2}, \frac{2\sigma_2^2}{n_2}\right),\label{eq:sample-joint-distribution}
\end{align}
where $\mathcal{N}(\cdot)$ denotes Gaussian distribution, $\mathcal{W}(\cdot)$ denotes Wishart distribution, and $\mathcal{X}^2(\cdot)$ denotes chi-squared distribution. 

Taking the derivative of the log of Eq.~\eqref{eq:sample-joint-distribution} with respect to our quantity of interest, $\mu_1$, we obtain
\begin{equation}\label{eq:log-likelihood-derivative}
    \frac{\partial \log(\mathcal{F})}{\partial \mu_1} = \frac{n_{1,2}(\bar{x}_1-\mu_1)}{\sigma_1^2(1-\rho_{1,2}^2)}+\frac{n_1(\bar{x}_1^\prime - \mu_1)}{\sigma_1^2}+\frac{n_{1,2}\rho_{1,2}(\bar{x}_2 - \mu_2)}{\sigma_1\sigma_2(1-\rho_{1,2}^2)}.
\end{equation}
Note that, in \texttt{Scenario 1}, $\mu_1$ is unknown while $\mu_2$, $\sigma_1$, $\sigma_2$, $\rho_{1,2}$ are known. Hence, \rev{we   set}  Eq.~\eqref{eq:log-likelihood-derivative}   equal to $0$, reorder the terms and obtain the maximum likelihood estimator for $\mu_1$ as:
\begin{align}
    \delta^{\text{MLE}} &= \frac{n_1(1-\rho_{1,2}^2)}{n_1(1-\rho_{1,2}^2)+n_{1,2}} \bar{x}_1^\prime +\notag\\
    &\qquad\frac{n_{1,2}}{n_1(1-\rho_{1,2}^2)+n_{1,2}}\left(\bar{x}_1 - \frac{\rho_{1,2}\sigma_1}{\sigma_2}(\bar{x}_2 - \mu_2)\right),
\end{align}
which is the same as the \rev{KFE} Eq.~\eqref{eq:kalman-uni-bi} that fuses estimators Eq. \eqref{eq:delta_1} and \eqref{eq:delta_12}  with optimal weights Eq.~\eqref{eq:g_1} and \eqref{eq:g_12} respectively.

\subsection{The MLE in \texttt{Scenario 2} Bivariate Case}\label{sec:likelihood-scenario2-bivariate}
As in the previous section, for deriving the MLE in this scenario we will also use the derivatives of the log of Eq.~\eqref{eq:sample-joint-distribution}.  However, since in this scenario  there are two unknown variables $\mu_1$ and $\mu_2$, apart from \rev{setting} Eq.~\eqref{eq:log-likelihood-derivative}  equal to $0$, we also need to account for the derivative of the log-likelihood with respect to $\mu_2$ and solve a system of two equations and two unknown variables.

\begin{equation}\label{eq:log-likelihood-derivative-2}
    \frac{\partial \log(\mathcal{F})}{\partial \mu_2} = \frac{n_{1,2}(\bar{x}_2 -\mu_2)}{\sigma_2^2(1-\rho_{1,2}^2)} + \frac{n_2(\bar{x}_2^\prime-\mu_2)}{\sigma_2^2} + \frac{n_{1,2}\rho_{1,2}(\bar{x}_1 - \mu_1)}{\sigma_1\sigma_2(1-\rho_{1,2}^2)}.
\end{equation}
By taking both Eq.~\eqref{eq:log-likelihood-derivative} and~\eqref{eq:log-likelihood-derivative-2} to equal to $0$ and solve the system of equations, we obtain
\begin{align}
      \hat{\mu}_1&=\frac{n_{1,2}n_1+n_1n_2(1-\rho_{1,2}^2)}{\Delta} \bar{x}_1^\prime +\notag\\
    &\,\,\frac{n_{1,2}^2 + n_{1,2}n_2}{\Delta}\left(\bar{x}_1 -\frac{n_{1,2}n_2}{n_{1,2}^2+n_{1,2}n_2}\frac{\rho_{1,2}\sigma_1}{\sigma_2}(\bar{x}_2^\prime - \bar{x}_2)\right),\\
    \hat{\mu}_2 &= \frac{n_{1,2}n_2+n_1n_2(1-\rho_{1,2}^2)}{\Delta} \bar{x}_2^\prime +\notag\\
    &\,\,\frac{n_{1,2}^2+n_{1,2}n_1}{\Delta}\left(\bar{x}_2 - \frac{n_{1,2}n_1}{n_{1,2}^2+n_{1,2}n_1}\frac{\rho_{1,2}\sigma_2}{\sigma_1}(\bar{x}_1^\prime - \bar{x}_1)\right),
\end{align}
where $\Delta = n_{1,2} n + n_1 n_2 (1-\rho_{1,2}^2)$
and $n=n_1+n_2+n_{1,2}.$
Then,
\begin{equation}
\delta^{\text{MLE}}=\hat{\mu}_1. 
\end{equation}

The variances of estimators $\hat{\mu}_1$ and $\hat{\mu}_2$ are given by
\begin{align}
    V(\hat{\mu}_1) &=  \frac{1+\beta(1-\rho^2)}{1+\alpha+\beta+\alpha \beta (1-\rho^2)}\frac{\sigma_1^2}{n_{1,2}} \\
& =\frac{(n_{1,2} + n_2(1-\rho^2))\sigma_1^2}{\Delta} \label{eq:var}\\
\textrm{V}(\hat{\mu}_2) &= \frac{1+\alpha(1-\rho^2)}{1+\alpha+\beta+\alpha \beta (1-\rho^2)}\frac{\sigma_2^2}{n_{1,2}}  \\
& =\frac{(n_{1,2} + n_1(1-\rho^2))\sigma_2^2}{\Delta} \label{eq:var2}
 \end{align}
 where
 \begin{equation}
    \rho=\rho_{1,2}, \quad \alpha=\frac{n_1}{n_{1,2}}, \quad \beta=\frac{n_2}{n_{1,2}}.
 \end{equation}

\begin{claim}[no opportunity for cooperation gains]
Given a budget of $n$ observations, the variance of $\hat{\mu}_1$ is minimized when all the $n$ observations are marginal observations of $X_1$, and the best unbiased estimator is the sample mean.
\end{claim}

\begin{proof}
Let  $(n_1, n_2, n_{1,2})$ denote a sample with $n_i$ marginal observations of $X_i$ and $n_{1,2}$ joint observations. Then, 
starting from sample   $(u,v,w)$, we can readily verify that variance is reduced if one can collect all marginal samples from $X_1$, and none from $X_2$, i.e.,  $(u+v,0,w)$ entails smaller variance than $(u,v,w)$. In addition, there is no gain by collecting joint samples, i.e.,  sample $(u+v+w-x,0,x)$ entails the same variance as  $(u+v,0,w)$, for any $0 \leq x \leq u+v+w$. Therefore, under the assumption that collecting local observations is always cheaper than collecting marginal observations from neighboring nodes or joint observations, we conclude that when means are unknown the sensors cannot benefit from cooperation. 

 We denote by  $V(\hat{\mu}_1;u,v,w)$ the variance of estimator $\hat{\mu}_1$ entailed by sample $(u,v,w)$. 
To show that $V(\hat{\mu}_1;u,v,w) > V(\hat{\mu}_1;u+v,0,w)$, note that
\begin{align}
V(\hat{\mu}_1;u,v,w) &= \frac{(1+v(1-\rho^2)/w)\sigma_1^2}{\Delta/w} \\
V(\hat{\mu}_1;u+v,0,w) &= \frac{\sigma_1^2}{w+u+v}
\end{align}
Letting $n=u+v+w$, we have
\begin{align}
& V(\hat{\mu}_1;u,v,w) - V(\hat{\mu}_1;u+v,0,w) = \\ 
&\quad =\frac{(1+v(1-\rho^2)/w)\sigma_1^2 n - \sigma_1^2 (n+ v(1-\rho^2) u/w)}{n\Delta} \nonumber \\
&\quad =\frac{ n -   u}{n\Delta/(\sigma_1^2 v(1-\rho^2)/w)}  > 0.
\end{align}

To show that $ V(\hat{\mu}_1;u+v,0,w)=V(\hat{\mu}_1;u+v+w-x,0,x)$, note that
\begin{align}
V(\hat{\mu}_1;u+v+w-x,0,x) &= \frac{ \sigma_1^2}{n}
\end{align}
which does not depend on $x$ and depends on $u$, $v$ and $w$ only through their sum, $n$. The above variance is attained by the sample mean. 
\end{proof}

\subsection{The MLE in \texttt{Scenario 2} Multivariate Case}\label{sec:likelihood-scenario2-multivariate}

Next, we show that, for $K$-variate Gaussians, the maximum likelihood estimators for mean parameters are the sample means even when joint observations are collected.

Denote a joint observation $(x_1, x_2,..., x_K)$ as $\mathbf{x}$.
The probability density function of  $K$-variate normal distribution is 
\begin{align}
   &\mathcal{N}(\mathbf{x}_{1},...,\mathbf{x}_{n}|{\pmb\mu}, \mathbf{\Sigma})\equiv  \nonumber \\
    &\prod_{i=1}^n (2\pi)^{\frac{-K}{2}} \det (\mathbf{\Sigma})^{\frac{-1}{2}} \exp\left\{-\frac{1}{2}(\mathbf{x}_{i}-{\pmb\mu})^\top \mathbf{\Sigma}^{-1} (\mathbf{x}_{i}-{\pmb\mu})\right\}.
    \end{align}
Hence, the log-likelihood function is 
\begin{align}
    &\mathcal{L}({\pmb\mu},  \nonumber \mathbf{\Sigma}|\mathbf{x}_{1},...\mathbf{x}_{n}) =-\frac{nk}{2}\ln (2\pi)-\\
    &-\frac{n}{2} \ln (\det (\mathbf{\Sigma})) - \frac{1}{2} \sum_{i=1}^{n}(\mathbf{x}_{i}-{\pmb\mu})^\top \mathbf{\Sigma}^{-1} (\mathbf{x}_{i}-{\pmb\mu}).
    \end{align}
Let $ \nabla_{\pmb\mu} \mathcal{L} ({\pmb\mu}, \mathbf{\Sigma}|\mathbf{x}_{1},...\mathbf{x}_{n}) $ be the gradient  of the log-likelihood. 
Then, we let  $ \nabla_{\pmb\mu} \mathcal{L} ({\pmb\mu}, \mathbf{\Sigma}|\mathbf{x}_{1},...\mathbf{x}_{n}) =0$ to find the maximum likelihood estimator for  ${\pmb\mu}$,
\begin{align*}
\nabla_{\pmb\mu}({\pmb\mu}, \mathbf{\Sigma}|\mathbf{x}_{1},...\mathbf{x}_{n}) & =- \sum_{i=1}^{n} \mathbf{\Sigma}^{-1} (\mathbf{x}_{i}\!-\!{\pmb\mu}) =\\
&= -\mathbf{\Sigma}^{-1} \sum_{i=1}^{n} (\mathbf{x}_{i}\!-\!{\pmb\mu}) = 0.
\end{align*}
The maximum likelihood estimator for ${\pmb\mu}$ is 
$$\hat{\pmb\mu} = \frac{1}{n}\sum_{i=1}^{n} \mathbf{x}_{i}.$$
The fact that the maximum likelihood estimators are the sample means shows that when all means are unknown and the covariance matrix is known, there is no advantage to collaborating across sensors for estimation purposes.


\begin{figure*}
\begin{minipage}{0.46\textwidth}
\begin{algorithm}[H]
\caption{ \texttt{DOUBLE-F} and \texttt{DOUBLE-Z}}\label{alg:DOUBLE-F-Z}
\begin{algorithmic}[1] 
\State \textbf{Initialize}: resource cost $\alpha\geq 0$, resource budget $E> 0$, no. of decision rounds $\mathcal{T} \equiv \lfloor TE/(\alpha+1)\rfloor$, no. of arms/variables $K$, exploration times $\mathcal{H}_\eta \equiv \{2K\} \cup \{\lceil \eta^\ell\rceil: \ell\in\mathbb{N}\}$, $\eta > 1$,   empirical surrogate reward $\hat{r}(j) \gets 0, \forall j \in [K]$ 

\For{$\tau = 1,...,\mathcal{T}$}
    \If{$\tau \leq 2K$} \Comment{Warm Up}
        \State Pull arm $J_{\tau} = ({\tau}\mod K)+1$
    \ElsIf{${\tau} \in \mathcal{H}_\eta$} \Comment{Exploration}
        \State Pull arm $J_{\tau}$ uniformly at random  from $[K]$
    \Else \Comment{Exploitation}
        \State \multiline{%
        Pull arm with largest empirical reward, i.e., \\ \qquad\qquad $J_{\tau} \gets \arg\max_{j\in [K]} \hat{r}(j)$}
    \EndIf
    \State CollectAndProcess($J_{\tau}$)
\EndFor
\end{algorithmic}
\end{algorithm}
\end{minipage}
\hfill
\begin{minipage}{0.46\textwidth}

\begin{algorithm}[H]
\caption{  \texttt{UCB-F} and \texttt{UCB-Z}}\label{alg:UCB-F-Z}
\begin{algorithmic}[1] 
\State \textbf{Initialize}: resource cost $\alpha\geq 0$, resource budget $E> 0$, no. of decision rounds $\mathcal{T} \equiv \lfloor TE/(\alpha+1) \rfloor$, no. of arms/variables $K$, \texttt{UCB} parameters $a$, $\epsilon = 1/\tau$, no. of pulls $n_{i, j} \gets 0$, $\forall j \in [K]$,  empirical surrogate reward $\hat{r}(j) \gets 0$, $\forall j \in [K]$

\For{$\tau = 1,...,\mathcal{T}$}
    \If{$\tau \leq 2K$} \Comment{Warm Up}
        \State Pull arm $J_{\tau} = ({\tau}\mod K)+1$
    \Else \Comment{Upper Confidence Bound}
        \State \multiline{%
        Construct confidence intervals $\texttt{CI}(j)$,  \\ \qquad\qquad  using Eq.~\eqref{eq:confidence-interval},  \quad $\forall j \in [K]$ }
        \State \multiline{%
        Pull the arm with highest UCB,  \\ \qquad\qquad  $J_{\tau} \gets \arg\max\limits_{j\in[K]} \hat{r}(j) + \texttt{CI}(j)$ }
    \EndIf
    \State CollectAndProcess($J_{\tau}$)
\EndFor
\end{algorithmic}
\end{algorithm}
\end{minipage}
\end{figure*}

\section{Pseudo Codes of Proposed Policies}\label{sec:pseudocode}
We present the pseudocode of \texttt{CollectAndProcess} in Algorithm~\ref{alg:collectandprocess}, that is used by \texttt{DOUBLE-F}, \texttt{DOUBLE-Z}, \texttt{UCB-F} and \texttt{UCB-Z}. The pseudocode of \texttt{DOUBLE-F} and \texttt{DOUBLE-Z} is shown in  Algorithm~\ref{alg:DOUBLE-F-Z} and the pseudocode of \texttt{UCB-F} and \texttt{UCB-Z} in Algorithm~\ref{alg:UCB-F-Z}.

\begin{algorithm}
\caption{ \texttt{CollectAndProcess}}\label{alg:collectandprocess}
\begin{algorithmic}[1] 
\State \textbf{Input}: {$J^\tau$}
        \If{$J^\tau \in \{2,  \ldots, K\}$ } \Comment{Collecting Bivariate Sample}
        \State Receive a joint sample $(x_1, x_{J^\tau})$ 
        \State Increase $n_{1, J^\tau}$ by $1$
        \If{under \texttt{UCB-F} or \texttt{DOUBLE-F}}
            \State \multiline{%
            Update surrogate reward of arm $J^{\tau}$, \\ \qquad\qquad using  Eq.~\eqref{eq:estimated-rho1} and \eqref{eq:estimated-rho}, \\ \qquad\qquad $\hat{r}(J^{\tau}) \gets\hat{\mathcal{I}}^{\tau}_{(X_1, X_{J^{\tau}})}(\mu_1)$ }
        \ElsIf{under \texttt{UCB-Z} or  \texttt{DOUBLE-Z}}
            \State \multiline{%
            Update surrogate reward of arm $J^{\tau}$, \\ \qquad\qquad  using Eq.~\eqref{eq:estimated-rho} and  \eqref{eq:z-trans-rho}, $\hat{r}(J^{\tau}) \gets z^{\tau}_{J^{\tau}}$} 
        \EndIf
    \Else \Comment{Collecting Univariate Sample(s)}
        \If{$E \geq \alpha+1$} 
            \State Receive a sample $x_1$
        \ElsIf{$1 \leq E < \alpha+1$}
            \State \multiline{%
            Receive $\lceil(\alpha+1)/E\rceil$ samples  \\ \qquad\qquad $x^{(1)}_1,...,x^{({\lceil(\alpha+1)/E\rceil})}_1$} 
        \ElsIf{$E < 1$}
            \State Receive $\lfloor(\alpha + 1)\rfloor$ samples $x^{(1)}_1,...,x^{({\lfloor(\alpha+1)\rfloor})}_1$ 
        \EndIf
        \If{under \texttt{UCB-F} or \texttt{DOUBLE-F}}
            \State Update surrogate reward of arm $1$, $\hat{r}(1) \gets 1$
        \ElsIf{under \texttt{UCB-Z} or \texttt{DOUBLE-Z}}
            \State \multiline{%
            Update surrogate reward of arm $1$, \\ \qquad\qquad  $\hat{r}(1) \gets \text{tanh}^{-1}(\sqrt{\alpha/(1+\alpha)})$ }
        \EndIf
    \EndIf
    \State Update Kalman filtering estimate of $\mu_1$ by Eq.~\eqref{eq:kalman-rho-unknown}  \\ \qquad\qquad \Comment{Update Estimate}
\end{algorithmic}
\end{algorithm}


\section{Applications}
Extensive sampling from the physical environment, e.g., air, water, and surface sampling, and from virtual ecosystems, e.g., network traffic, and collaboratively learning a characteristic parameter about the environment is a landmark of modern computer and communication systems~\cite{romer2004design, akyildiz2002wireless}.
Its applications range from smart home monitoring and military coalition to environmental monitoring, where characteristic parameters may be temperature, GPS-related signals, or air pollution, respectively~\cite{zhao2003collaborative}.
In the following, we describe two applications where our model and analysis can be applied.

\emph{\textbf{IoT DDoS Attack Detection.}}
As the number of devices connecting to home networks, e.g., PCs, tablets, mobile devices, and IoT devices like smart thermostats, keeps increasing in recent years, it has attracted the attention of malicious agents interested in compromising those devices and launching distributed denial of service (DDoS) attacks~\cite{marzano2018evolution}. Many Internet service providers have installed software at home routers that are used to periodically make a variety of observations such as numbers of packets and bytes uploaded and downloaded. These observations can be used to estimate their means and/or correlations. One can model this as a collection of multimodal sensors in a home router and/or a set of sensors of the same modality distributed across homes.  Data from these sensors are then collected at a data center subject to constraints on available bandwidth from the home router to the data center.  Such a design has been used to develop detectors for DDoS attacks~\cite{mendoncca2019extremely}.

\emph{\textbf{Distributed Estimation in Wireless Sensor Network.}}  
In wireless sensor networks (WSNs), 
communication consumes the most energy, which is the critical resource in systems with limited on-board batteries
~\cite{zhao2003collaborative}, posing the challenges of determining whether devices should collaborate or not, and setting the rate at which information must be transmitted through the network given the metrics to be estimated. 
Sensors in WSNs map naturally to sensors/agents in our model.
If sensors all sense the same variable, there are usually certain spatial correlations between observations~\cite{vuran2006spatial}; if the controller of the sensors has sufficient prior knowledge about the correlation structure, learning tasks map to our \texttt{Scenario 1} or \texttt{Scenario 2}. If sensors sense different modalities~\cite{zhao2003collaborative} whose correlation structure may not be obvious for the controller, our results about \texttt{Scenario 3} can be applied.


\section{Table of Notation}\label{sec:notation}

Table~\ref{tab:notation} summarizes the notation used throughout this work. 

\begin{table}[t]
    \centering
        \caption{Summary of notation. }
    \label{tab:notation}
    \myhl{
    \scriptsize
    \begin{tabular}[P]{|c|l|}
        \hline
         \textbf{Notation} & \textbf{Description} \\
         \hline\hline
         $K$ & Number of sensors/agents \\
         $S_k$ & Sensor/agent $k$, $k \in [K] \equiv \{1,...,K\}$\\
         $X_k$ & Random variable characterizing  observations from sensor/agent $k$ \\
         $x_{k, i}$ & $i$-th observation from sensor/agent $k$, where $X_{k,i} \sim X_k$\\
         $\mu_k$ & Mean of $X_k$, $\mu_k=\mathbb{E}[X_k]$ \\
         $\rho_{k, \ell}$ & Pearson correlation coefficients of $X_k$ and $X_\ell$ \\
         $\sigma_k^2$ & Variance of $X_k$\\
         ${\bm \mu}$ & Mean vector, ${\bm \mu}=(\mu_1, ...\mu_K)$\\
         ${\bm \Sigma}$ & Covariance matrix whose $(k, l)$-th entry $({\bm \Sigma})^{k, \ell} = \rho_{k, \ell}\sigma_k\sigma_\ell$\\
         \hline
         \hline
         $\alpha$ & Communication cost per observation\\
         $E$ & Resource budget per sensor/agent per time slot\\
         $\text{Pw}([K])$ & Power set of $[K] \equiv \{1, 2, ..., K\}$\\
         $p_{\mathcal{K}}$& Prob. that  sensors $S_k$, for $k\in \mathcal{K}$, are active;  others are inactive\\
        $\nobs$ & Random variable characterizing number of all observations \\      $\realizationnobs$ & Actual number of all observations \\    
        $N_{\mathcal{K}}$ & Random variable characterizing number of joint   \\
        & observations collected from   $\mathcal{K}$, $\mathbb{E}[N_{\mathcal{K}}] = p_{\mathcal{K}} \realizationnobs$ \\
        $n_{\mathcal{K}}$ & Actual   number of joint observations collected from  $\mathcal{K}$ \\ 
        \hline
        \hline
        $T$ & Total number of time slots \\ 
        $t$ & A time slot\\
         $\tau$ & A decision round in the MAB model, see Table~\ref{tab:mabresconst}\\
         $\mathcal{T}$ & Total number of decision rounds in the MAB model, $\mathcal{T} \leq T$\\
         $J_\tau$ & Arm selected by MAB learner in round $\tau$\\
         \hline
         \hline \rule{0pt}{1.05\normalbaselineskip}          $\bar{X}_k$ & Sample mean of $X_k$\\
         $\delta_{\mathcal{K}}$ & Estimator that depends on samples of $(X_k: k\in \mathcal{K})$\\
         $\delta^*$ & Kalman filter estimator \rev{(KFE)}\\
         $g_{\mathcal{K}}$ & Weight corresponding to $\delta_{\mathcal{K}}$ in Kalman filter\\
         $\mathcal{I}_{X_k}(\mu_\ell)$ & Fisher information w.r.t. $\mu_l$ conveyed by an observation of $X_k$\\
         $\hat{\mathcal{I}}_{X_k}(\mu_\ell)$ & Estimated Fisher information by samples observed up until round $\tau$\\
         $\hat{\bm \Sigma}$& Estimated covariance matrix by samples observed up until round $\tau$\\
         $\hat{\rho}$ & Estimated correlation coeff. by samples observed up until round $\tau$\\
         $z$ & $z$-transformed correlation coefficient estimate\\
         \hline
    \end{tabular}
    }
\end{table}

\bibliography{ref}
\bibliographystyle{IEEEtran}

\begin{IEEEbiographynophoto}{Yu-Zhen Janice Chen} is currently a Ph.D. candidate at the University of Massachusetts, Amherst. She received her BS.c. Degree in computer science from the Chinese University of Hong Kong in 2019.  Her research interests include statistical machine learning, sequential decision-making, performance analysis, modeling, and algorithm design for computing systems. 
\end{IEEEbiographynophoto}
\begin{IEEEbiographynophoto}{Daniel S. Menasch\'e}
received the Ph.D. degree
in Computer Science from the University of Massachusetts, Amherst, in 2011. Currently, he is an associate professor at the Institute of Computing, Federal University of Rio de Janeiro, Brazil.
His interests are in modeling, analysis, security, and
performance evaluation of computer systems, being
a recipient of best paper awards at GLOBECOM
2007, CoNEXT 2009 and  INFOCOM 2013. He is an alumni-affiliated member of the
Brazilian Academy of Sciences.
\end{IEEEbiographynophoto}
\begin{IEEEbiographynophoto}{Don Towsley}
(Fellow, IEEE and ACM) holds a
Ph.D. in Computer Science (1975) from the University of Texas. He is currently a Distinguished
Professor at the Manning College of Information
\& Computer Sciences, University of Massachusetts, Amherst. His research interests
include performance modeling and analysis, as well as
quantum networking. He has received several
achievement awards, including the 2007 IEEE
Koji Kobayashi Award and the 2011 INFOCOM
Achievement Award.
\end{IEEEbiographynophoto}

\vfill

\end{document}